\documentclass{article}
\usepackage{ijcai21}

\usepackage{times}

\usepackage{soul}
\usepackage{url}
\usepackage[hidelinks]{hyperref}
\usepackage[utf8]{inputenc}
\usepackage[small]{caption}
\usepackage{graphicx}
\usepackage{amsmath}
\usepackage{booktabs}
\usepackage{graphics}
\usepackage{makecell}
\usepackage{diagbox}
\usepackage{multirow}
\usepackage{placeins} 
\usepackage{appendix}
\usepackage{longtable} 

\usepackage{url,hyperref,lineno,microtype,subcaption}
\usepackage[onehalfspacing]{setspace}
\usepackage{subcaption}
\usepackage{float}

\usepackage{xcolor}

\definecolor{myGreen}{rgb}{0.25,0.60,0.15}
\definecolor{myRed}{rgb}{0.85,0.40,0.30}

\newcommand{\green}[1]{{\color{myGreen}#1}}
\newcommand{\red}[1]{{\color{myRed}#1}}

\title{Plotting time: On the usage of CNNs for time series classification}

 \author{
 Nuno M. Rodrigues$^1$\and
 João E. Batista$^1$\and
 Leonardo Trujillo$^{2}$\And
 Bernardo Duarte$^{3,4}$ \\ \And
 Mario Giacobini$^5$\And
 Leonardo Vanneschi$^6$\And
 Sara Silva$^1$
 \affiliations
 $^1$LASIGE, Faculdade de Ciências da Universidade de Lisboa, Campo Grande, 1749-016 Lisboa, Portugal\\
 $^2$Departamento de Ingeniería Eléctrica y Electrónica, Tecnológico Nacional, México/IT de Tijuana\\
 $^3$MARE - Marine and Environmental Sciences Centre, Faculdade de Ciências da Universidade de Lisboa, Campo Grande, 1749-016 Lisboa, Portugal\\
 $^4$Departamento de Biologia Vegetal, Faculdade de Ciências da Universidade de Lisboa, Campo Grande, 1749-016 Lisboa, Portugal\\
 $^5$Data Analysis and Modeling Unit, Department of Veterinary Sciences, University of Torino, Italy\\
 $^6$NOVA Information Management School (NOVA IMS), Universidade Nova de Lisboa, Campus de Campolide, 1070-312 Lisboa, Portugal
 \emails
 \{nmrodrigues, jebatista, baduarte, sara\}@fc.ul.pt,
 leonardo.trujillo@tectijuana.edu.mx,
 mario.giacobini@unito.it,
 lvanneschi@novaims.unl.pt
 }

\begin{document}

\maketitle

\begin{abstract}

We present a novel approach for time series classification where we represent time series data as plot images and feed them to a simple CNN, outperforming several state-of-the-art methods.

We propose a simple and highly replicable way of plotting the time series, and feed these images as input to a non-optimized shallow CNN, without any normalization or residual connections. These representations are no more than default line plots using the time series data, where the only pre-processing applied is to reduce the number of white pixels in the image. We compare our method with different state-of-the-art methods specialized in time series classification on two real-world non public datasets, as well as 98 datasets of the UCR dataset collection. The results show that our approach is very promising, achieving the best results on both real-world datasets and matching~/~beating the best state-of-the-art methods in six UCR datasets. We argue that, if a simple naive design like ours can obtain such good results, it is worth further exploring the capabilities of using image representation of time series data, along with more powerful CNNs, for classification and other related tasks.

\end{abstract}

\section{Introduction}

Time series classification~(TSC) problems are highly diverse in nature, covering a wide range of domains. Recently, this field of study is receiving more and more attention, arguably due to the creation of large data repositories such as UCR~\cite{UCRArchive2018}. Such easy availability of data motivated the proposal of multiple new algorithms to address TSC.

When looking at the current state-of-the-art~(SOTA) of TSC, led by the HIVE-COTE or TS-CHIEF methods, or even just the more popular algorithms for TSC, there are two main approaches. Some algorithms like BOSS, BOP or WEASEL use word and symbol embeddings along with sliding windows to transform the data~\cite{Schfer2014TheBI,Large2018FromBT,Schfer2017FastAA}. The other predominant approach is to use different ensemble methods, such as HIVE-COTE (which includes BOSS), TS-CHIEF or Proximity Forests \cite{Lines2016HIVECOTETH,Lucas2019ProximityFA,Shifaz2020TSCHIEFAS}.

Recently, a new trend is emerging due to the rising popularity of convolutional neural networks~(CNN), which is the use of convolution operations. We can see this in ROCKET~\cite{Dempster2020ROCKETEF}, which applies random convolutional kernels to the time series data, or in InceptionTime, which is an actual computer vision model \cite{L7,Fawaz2020InceptionTimeFA}.

The vast majority of these algorithms share one common trait: they use the numeric data of the time series to find patterns and induce models.
However, pattern recognition can also be tackled from a computer vision perspective, using visual representations of the data. As previously mentioned, and further elaborated, there are already some recent algorithms that use a visual representation of the data to perform TSC. However, we argue that these works follow an overly complex path, and we propose a much simpler approach.

Our work demonstrates that, with highly simplistic visual representations of the time series data, it is possible to achieve results that are equal or even better than the ones obtained by the best SOTA methods available today.



\section{Related work}

There are two general goals when non-image data is converted to a 2D image representation.
The first one is to understand and interpret the data, since converting it to a visual representation allows a researcher
to analyze the data in a more human-readable form.
These techniques have a long history in pattern analysis and machine learning \cite{L1},
including different types of geometric representations, pixel-oriented visualizations and dimensionality reduction techniques.
For instance, a popular SOTA method is t-SNE \cite{L2}, that allows for 2D and 3D representations of large datasets in high-dimensional spaces.
While t-SNE focuses on visualizing an entire dataset, General Line Coordinates (GLC), on the other hand, can be used to represent individual data samples as polylines \cite{L3},
without any loss of information.

The second broad reason for transforming non-image data into images is that it allows researchers
to solve the original learning problem using powerful and widely available computer vision algorithms, such as CNNs.
This approach is much more recent, with a growing but still fairly small literature on the subject.
In \cite{L4} the authors used GLC representations of data to pose an image recognition problem and solved it with CNNs,
achieving similar performance compared to solving the original problem directly, with the added benefit that interpreting the results is easier.
RNA data is transformed into 2D images in \cite{L5}, outperforming previous works on the same problem, while using a domain specific transformation of the data.
A notable approach is DeepInsight \cite{L6}, that proposes a general methodology to transform data samples into unique images and subsequently
applies CNN to a variety of problems from different domains.
DeepInsight transforms the data by applying Principal Component Analysis in a simple but creative manner, and also produces
informative visualizations of the samples in a dataset, intended to provide unique \textit{insights} on the problem.
Regarding the case of time series classification, \cite{L7} transforms the time series to a Gramian Angular Difference Field (GADF),
a visual representation of the data that is subsequently processed
by Google's pre-trained Inception CNN, achieving SOTA results.


\section{Proposed methodology}

\begin{figure}[]
\includegraphics[width=\linewidth]{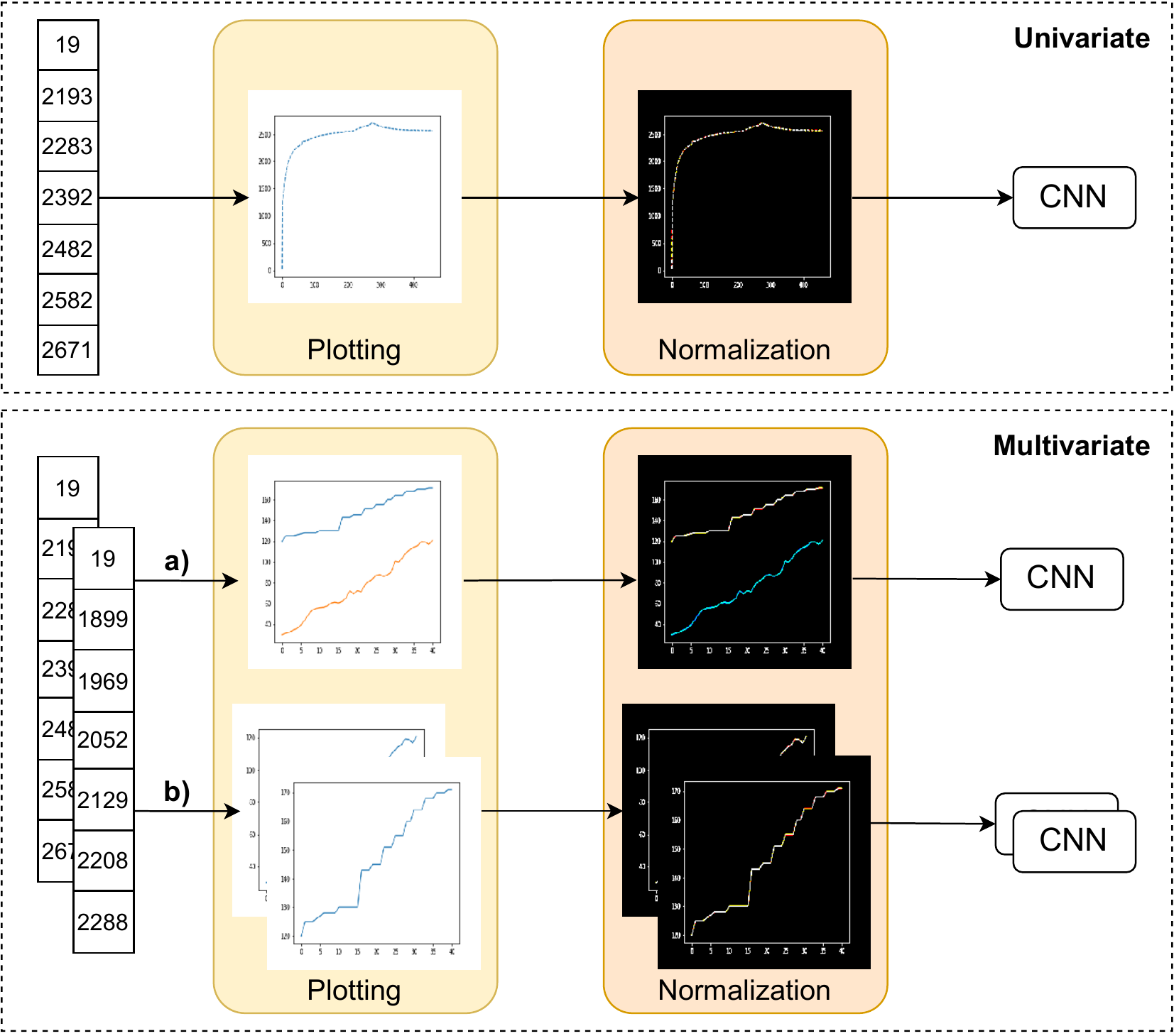}
\caption{The proposed methodology. For univariate time series, each series is converted into an image, normalized and given to a CNN. For multivariate time series, we test two approaches: \textbf{a)} all series are displayed on the same plot and given to a single CNN; \textbf{b)} different series are displayed in separate plots and given to siamese CNNs, whose output is concatenated before flattening. We also split the multivariate cases in multiple univariate problems, for comparison.
}
\label{fig:methodology}
\end{figure}

The methodology we propose is more focused on the transformation of the time series data into image data, and less on the particulars of the CNN architecture. Figure~\ref{fig:methodology} illustrates the different steps of the methodology, in both univariate and multivariate cases. Details of each step are provided below.

\subsection{Generating image data from time series}

We follow a simple and replicable process of transforming the time series data into image plots to be used by the CNN.
First, the time series data is loaded without any pre-processing.
Then, using \textit{matplotlib} library~\cite{Hunter:2007} we plot each time series adopting the default parameters, obtaining a $432 \times 288$ sized image for each series, as seen in Figure~\ref{fig:methodology}. 
The plot axes are included in the image, along with their problem specific time ($x$-axis) and value ($y$-axis) numbers, since they can provide important information, as shown later when we analyse feature maps.
Notice that not all the time series are required to have the same length, since this has no impact when transforming the data into an image, and does not imply loss of valuable information~(this way avoiding the need to oversample the data).

\subsection{Normalizing the image data}
\label{subsection:normalizing_image_data}
When loading the plots to use as input to our networks, we perform two kinds of image pre-processing.
We start by the standard rescaling of pixel values to the $[0,1]$ range, and then we apply a samplewise standard normalization.
The samplewise normalization essentially inverts the colors~(produces a negative of the original image, as shown in Figure~\ref{fig:methodology}), which reduces the amount of empty white space by turning the pixel values to 0 (black). This reduces the activations of the feature maps in empty areas where there is no information, while increasing them in the numbers of the axes, which is helpful as they also contain information. This normalization produces large improvements in both learning and generalization.
Its effect on the feature maps will be analysed in Section~\ref{subsection:contribution_normalization}.

\subsection{Feeding the plots to the CNN}
\label{subsection:modeling}

For all the problems addressed, we use a simple shallow CNN of five convolutional and pooling layers, interspersed, and three fully connected layers with two dropouts in between.
As for the hyperparameters, we use standard values and do not perform any sort of grid-search / bayesian hyperparameter optimization.
Regarding the filters, we start with $16$ and double the number in each convolutional layer. In the first and second fully connected layers we use $256$ and $128$ units, respectively.
For the optimizer we use ADAM with a learning rate of $0.001$.

When modeling a problem, if its data is univariate (containing only one time series) we simply feed the pre-processed images to a CNN like the one described above (top part of Figure~\ref{fig:methodology}). For multivariate problems (data containing more than one time series) we test different approaches, identified as a) and b) in the bottom half of Figure~\ref{fig:methodology}.
In variant a) the different time series for the same sample are displayed on the same plot and given to a single CNN as the one used for the univariate approach. In variant b) the different time series are displayed in separate plots and given to a set of siamese CNNs. Each input head has the same set of convolutional and pooling layers described above and, before flattening, the outputs of these networks are concatenated and fed into the same classifier section, producing a single output. This method allows for different features to be extracted from the different time series, however, it is far more computationally expensive.

There is also an additional variant, not shown in the figure, which is to split the multivariate problems in multiple univariate problems, each one using only one time series. We have included this last variant because we wanted to check that drawing more than one time series in a single plot, or using a set of siamese networks, is not a confounding factor that prevents the learning of the information contained in each variable. Since our multivariate datasets have only two time series, it was feasible to do so.


\section{Experiments}

All experiments were performed on a Windows 10 machine with one NVIDIA 2080 TI GPU with 11GB of RAM. All network based models were implemented in TensorFlow.

\subsection{Datasets and Problems}

Here we describe the datasets used to test our methodology.
Since the UCR dataset is public domain and is already described in its website, we provide more details on the remaining two datasets and respective prediction tasks.

\subsubsection{OPTOX}
The OPTOX dataset was collected by the Marine and Environmental Sciences Centre - MARE - Faculty of Sciences, University of Lisbon. 
Each sample corresponds to a chlorophyll fluorescence induction curve, and contains the fluorescence values taken at different time steps from a model diatom used in ecotoxicology~(\textit{Phaeodactylum tricornutum}) exposed to 13 different emerging contaminants at different concentrations, following the international standards for ecotoxicological assays. For each concentration, 30 independent replicates were obtained. Details of the experiments and samples can be found in~\cite{Silva2020ComfortablyNE}.

With this dataset, we address the challenge of identifying which of the 13 contaminants is detected, using only the fluorescence induction curve.
Table~\ref{table:datasetOPTOX} contains more detailed information regarding this dataset, including the identification of the contaminants, number of concentrations considered for each, and number of samples per class.

\begin{table}[]
\centering
\caption{OPTOX dataset summary, containing the class names (contaminants) concentrations considered for each contaminant, and distribution of the total number of samples per class.}
\scalebox{0.75}{
\begin{tabular}{c|ccc}

Contaminant & \#concentrations  & \#samples\\
\hline
Diclofenac     & $6$&   $180$\\
Fluoxetine      & $6$&   $180$\\
Glyphosate   & $6$&   $180$\\
Ibuprofen     & $6$&   $180$\\
Propranolol    & $6$&   $180$\\
Sodium Dodecyl Sulphate (SDS)     & $5$&   $150$\\
Triclosan     & $6$&   $180$\\
Dissolved Ionic Copper        & 4  & 120\\
Copper Engineered Nanoparticles       & 4  & 120\\
Dissolved Ionic Zinc        & 4  & 120\\
Zinc Engineered Nanoparticles       & 4  & 120\\
Dissolved Ionic Titanium        & 4  & 120\\
Titanium Engineered Nanoparticles       & 4  & 120\\
\hline

\end{tabular}
\label{table:datasetOPTOX}
}
\end{table}

\subsubsection{FISIO}
The FISIO dataset was collected by the Centre of Preventive Medicine and Sport - SUISM - University Structure of Hygiene and Sport Sciences, Centre of excellence of the University of Turin. Data were recorded during an indoor trial conducted by the centre, where a group of 262 volunteers underwent an aerobic exercise on a trade mill. The test started with a speed of 5 km/h that was incremented by 1 km/h every minute. The participant could suspend the test when he/she felt exhausted. Cardio-respiratory variables were recorded every 10 seconds, namely heart rate (HR) and ventilation (VEN), along with the age, gender and body mass index (BMI) of each participant. More details on this dataset and related tests can be found in~\cite{Azzali2020TowardsTU}.

With this dataset, we addressed two challenges: based on both HR and VEN, we predict 1) BMI class: two classes were defined, based on the UK NHS BMI charts for both children and adults (since some of the participants were underage), where Class 1 contains Underweight and Healthy participants, and Class 2 contains Overweight and Obese participants; 2) Age class: two classes were defined, with Class 1 for participants younger than 30 years old and Class 2 for all the others. Gender remained unused.

Table~\ref{table:datasetFISIO} contains the information regarding sample distribution per class for both problems.
Since we had both time series (HR and VEN) for each participant, we decided to use both multivariate and univariate approaches (Section~\ref{subsection:modeling}).

\begin{table}[]
\centering
\caption{FISIO dataset summary, containing the distributions of samples per class for the two problems addressed, BMI and Age.}
\label{table:datasetFISIO}
\scalebox{0.75}{
\begin{tabular}{ccccc}
\hline
\multicolumn{2}{c}{BMI} &  & \multicolumn{2}{c}{Age} \\ \cline{1-2} \cline{4-5}
Class 1    & Class 2    &    & Class 1    & Class 2    \\ \cline{1-2} \cline{4-5}
221        & 41    &   & 128        & 134        \\ \hline
\end{tabular}
}
\end{table}

\subsubsection{UCR}
The UCR archive contains, as of now, a set of 128 univariate and 30 multivariate datasets~\cite{UCRArchive2018}.
From the 128 univariate problems we tested our approach on 98 of them.
The selection criterion for these 98 problems was two-fold. First, because the results provided by UCR for each classifier do not include all 128 problems, we decided to use only the problems that were tested with all the algorithms, ending up with 108 problems.
Second, due to time constrains, we decided to not test our approach on single axis data of multi-axes problems~(\textit{e.g} Cricket X/Y/Z), ending with the final amount of 98 problems.
Regarding the multivariate problems, the archive does not have results to compare to, so we did not use them.

\subsection{Methods}
\label{subsection:methods}

In order to have comparison grounds for our methodology, we tested a set of different classifiers for both the OPTOX and the FISIO datasets: XGBoost~\cite{2016XGBoostAS}, Random Forests~(RF)~\cite{Breiman2001}, deep learning classifiers 1D CNN and a deep ANN, and ROCKET.

The FISIO dataset presented an additional problem for these classifiers, as it consists of a multivariate problem with time series of different lengths. To address this issue, we truncated each time series to the same length as the smallest one of that type, therefore assuming the risk of losing information. 
This problem serves to highlight the advantage of our methodology in being able to use the full time series without the need for all the series to have the same length.

We performed 30 independent runs with all the methods, with an 80/20 split for training and test, each with a seed equal to the number of the run.
For each run, both XGBoost and RF were optimized by means of grid-search for the following parameters: max depth; number of estimators; tree method; learning rate; criterion.

As for UCR, the archive already provides results based on 30 independent runs from several other algorithms.
Due to the limited computational resources and large amount of problems, we started by performing five independent runs per problem, and an additional 25 only in those problems where our methodology was able to outperform or tie as the best result.


\section{Results and Discussion}

Here we compare the results obtained by our approach with the results of several other methods, including the current SOTA, on the OPTOX, FISIO and UCR datasets. Statistically significant differences in the results are determined by the Kruskal-Wallis test at $p<0.01$. We use a non-parametric test because not all the results follow a normal distribution. When the results presented in tables identify more than one value as the best, it means the difference between them is not statistically significant. All the $p$-values are included in the Appendix, in Table~\ref{pvalues_xeno} for OPTOX, Tables~\ref{pvalues_BMI_Siamese} to~\ref{pvalues_Age_HR+VEN} for FISIO, and Tables~\ref{pvalues_bme} to~\ref{pvalues_trace} for the UCR problems where we performed 30 runs (see Section~\ref{subsection:methods}).  
This section also discusses the effect that image data normalization has on the learned features by examining feature maps extracted from the CNNs. 
\subsection*{OPTOX}

For this dataset, we have produced two sets of results from our methodology, identified as CNN and CNN Log. 
In CNN, the values of the different time steps are drawn at equally spaced intervals on the $x$-axis of the plots, even if in reality the time intervals between every two measurements are not the same for the entire curve. By convention, when visualizing an induction curve it is common to use a logarithmic scale on the $x$-axis, which is what CNN Log does.

Table~\ref{results:optox} compares all tested classifiers in terms of median accuracy in both training and test sets for the contaminant prediction task of the OPTOX dataset. As shown, our algorithm outperformed all the other methods. CNN Log was the best performing method, significantly better than CNN, which is tied in second with ROCKET. This difference between CNN and CNN Log highlights the need to take into consideration domain knowledge when transforming the data into images. Despite not changing the methodology whatsoever, this change in scale significantly improved the results.

\begin{table}[]
\centering
\caption{Median overall accuracy of all methods on the OPTOX problem. The best test result is represented in \green{\underline{green}}.}
\resizebox{\linewidth}{!}{
\label{results:optox}


\begin{tabular}{l|ccccccc}

	 & RF 		& XGBoost & ROCKET & ANN & 1D CNN & CNN Log & CNN \\
	 \hline
Train&$0.9625$	& $0.9983$ &$0.9991$ &$0.8493$ &$0.9993$ &$0.9807$ &$0.9815$ \\
Test &$0.7461$	& $0.8076$&$0.9705$ &$0.9089$ &$0.9641$ & \green{\underline{$0.9765$}}&$0.9732$ \\
\end{tabular}

}
\end{table}

\subsection*{FISIO}

Table~\ref{results:dati} compares all tested classifiers in terms of median accuracy in both training and test sets, for the BMI and Age problems of the FISIO dataset. The results identified as HR and VEN are the ones obtained using only one of the time series (the additional variant mentioned in Section~\ref{subsection:modeling}). The results identified as HR+VEN refer to variant a), where both time series are drawn in the same plot, and the ones identified as Siamese refer to variant b), where each time series is drawn in a different plot and the two plots are given to siamese networks. 

On the BMI classification problem, the CNNs using only the HR time series obtained the best median result, but not significantly different from the results of RF and XGBoost. This provides two interesting findings: for a multivariate problem, the best results were obtained using only one of the time series, supporting the hypothesis that using multiple time series may indeed be a confounding factor (for 1D CNN, HR+VEN certainly is); two standard classification methods were able to outperform not only all other deep learning classifiers, but also ROCKET, which is one of the top algorithms for TSC.

Regarding the Age classification problem, the best median accuracy was obtained by the CNNs, who achieved the exact same value using only the HR time series, using both HR+VEN on the same plot, and following the siamese approach.
When using only the HR time series, the CNNs outperformed all other methods, which did not happen with the other two approaches where, despite having a higher median accuracy, CNNs were not significantly different from the other highlighted classifiers.


\begin{table}[]
\centering
\caption{Median overall accuracy from all models in the BMI and Age problems of the FISIO dataset. The best test result for each problem is represented in \green{\underline{green}}.}
\label{results:dati}
\scalebox{0.65}{
\begin{tabular}{lcccccccc}
\hline
               & \multicolumn{8}{c}{BMI}                                                                                     \\ \cline{2-9} 
               & \multicolumn{2}{c}{HR} & \multicolumn{2}{c}{VEN} & \multicolumn{2}{c}{HR+VEN} & \multicolumn{2}{c}{Siamese} \\ \cline{2-9} 
               & Train      & Test      & Train      & Test       & Train        & Test        & Train        & Test         \\ \hline
RF & 0.8564     & \green{\underline{0.8301}}    & 0.9090     & 0.8113     & 0.9138       & 0.8113      & ----         & ----         \\
XGBoost        & 0.8562     & \green{\underline{0.8301}}    & 0.9371     & 0.8301     & 0.9251       & 0.8301      & ----         & ----         \\
ROCKET            & 1          & 0.7264    & 1          & 0.7547     & 1            & 0.7169      & ----         & ----         \\
ANN         & 0.9330     & 0.7641    & 0.9617     & 0.7924     & 0.9760       & 0.7735      & 0.8492       & 0.8113       \\
1D CNN         & 1          & 0.6792    & 1          & 0.6981     & 0.5100       & 0.46        & 1            & 0.7547       \\
CNN            & 1          & \green{\underline{0.8461}}    & 1          & 0.8269     & 1            & 0.8269      & 1            & 0.8365      
\end{tabular}
}
\scalebox{0.65}{
\begin{tabular}{lcccccccc}
\hline
               & \multicolumn{8}{c}{Age}                                                                                     \\ \cline{2-9} 
               & \multicolumn{2}{c}{HR} & \multicolumn{2}{c}{VEN} & \multicolumn{2}{c}{HR+VEN} & \multicolumn{2}{c}{Siamese} \\ \cline{2-9} 
               & Train      & Test      & Train      & Test       & Train        & Test        & Train        & Test         \\ \hline
RF & 0.9569     & 0.6037    & 0.9377     & 0.6037     & 0.9808       & 0.6415      & ----         & ----         \\
XGBoost        & 0.9730     & 0.6037    & 0.9491     & 0.6320     & 1            & \green{\underline{0.6792}}      & ----         & ----         \\
ROCKET            & 1          & 0.6132    & 1          & 0.5849     & 1            & \green{\underline{0.6981}}      & ----         & ----         \\
ANN         & 0.8947     & 0.6415    & 0.9282     & 0.6037     & 0.9665       & \green{\underline{0.6981}}      & 0.7775       & \green{\underline{0.6981}}       \\
1D CNN         & 1          & 0.5660    & 1          & 0.5849     & 1            & 0.6698      & 0.9952       & \green{\underline{0.6792}}       \\
CNN            & 1          & \green{\underline{0.7058}}    & 1          & 0.6568     & 1            & \green{\underline{0.7058}}      & 1            & \green{\underline{0.7058}}      
\end{tabular}
}
\end{table}

\subsection*{UCR}
\label{subsec:ucr}

Table~\ref{results:ucrrank} compares the number of times that each method, both the ones provided by UCR and ours, achieved first, second or third ranking on the set of 98 problems addressed.
We can see that, although it is clearly not the best, our approach is able to challenge the current SOTA methods, beating~/~matching them on six problems, and arriving in third on another.

As mentioned in Section~\ref{subsection:methods}, in order to assess the results on such a large number of datasets, we performed five independent runs on all addressed problems, and then compared the best test accuracy achieved on each problem (between our five runs and UCR's reported 30 runs per method). 
The six problems where CNN matched/beat the other methods on this comparison are BME, Coffee, Earthquakes, Plane, SmoothSubspace and Trace~(see Appendix, Table~\ref{giga_table}). For these six problems, we performed our additional 25 independent runs, and compared the median test accuracy of the 30 runs with the median results provided by UCR. As shown in Table~\ref{results:ucrmedians}, CNN achieved the best results in all these problems.

\begin{table*}[]
\centering
\caption{Number of times each method achieved first, second or third ranking spot in the UCR problems.} 
\resizebox{\linewidth}{!}{
\begin{tabular}{c|ccccccccccccccc}
&CNN&TS-CHIEF&HIVE-COTE&ROCKET&InceptionTime&STC&ResNet&ProximityForest&WEASEL&S-BOSS&cBOSS&BOSS&RISE&TSF&Catch22\\ 
\hline
First&6&35&32&26&31&21&25&23&20&20&20&18&12&14&11\\ 
Second&0&6&9&10&12&2&9&4&7&4&2&4&1&2&1\\ 
Third&1&7&5&12&5&6&11&4&5&4&3&3&0&1&5\\ 
\end{tabular}
}
\label{results:ucrrank}
\end{table*}

\begin{table*}
\centering
\caption{Median overall test accuracy in the selected UCR problems. The best results for each problem are highlighted in \green{\underline{green}}.}
\resizebox{\linewidth}{!}{
\begin{tabular}{l|cccccccccccccccc}
&CNN&TS-CHIEF&HIVE-COTE&ROCKET&\makecell{Inception\\Time}&STC&ResNet&\makecell{Proximity\\Forest}&WEASEL&S-BOSS&cBOSS&BOSS&RISE&TSF&Catch22\\
\hline
BME&\green{\underline{$1$}}&\green{\underline{$1$}}&$0.9867$&\green{\underline{$1$}}&\green{\underline{$1$}}&$0.9400$&\green{\underline{$1$}}&\green{\underline{$1$}}&$0.9600$&$0.876$&$0.7733$&$0.8667$&$0.7933$&$0.9733$&$0.9133$\\
Coffee&\green{\underline{$1$}}&\green{\underline{$1$}}&\green{\underline{$1$}}&\green{\underline{$1$}}&\green{\underline{$1$}}&\green{\underline{$1$}}&\green{\underline{$1$}}&\green{\underline{$1$}}&\green{\underline{$1$}}&\green{\underline{$1$}}&\green{\underline{$1$}}&\green{\underline{$1$}}&\green{\underline{$1$}}&\green{\underline{$1$}}&\green{\underline{$1$}}\\
Earthquakes&\green{\underline{$0.7698$}}&$0.7482$&$0.7482$&$0.7482$&$0.7410$&$0.7410$&$0.7194$&$0.7518$&$0.7482$&$0.7482$&$0.7482$&$0.7482$&$0.7482$&$0.7482$&$0.7410$\\
Plane&\green{\underline{$1$}}&\green{\underline{$1$}}&\green{\underline{$1$}}&\green{\underline{$1$}}&\green{\underline{$1$}}&\green{\underline{$1$}}&\green{\underline{$1$}}&\green{\underline{$1$}}&\green{\underline{$1$}}&\green{\underline{$1$}}&\green{\underline{$1$}}&\green{\underline{$1$}}&\green{\underline{$1$}}&\green{\underline{$1$}}&$0.9905$\\
SmoothSubspace&\green{\underline{$1$}}&\green{\underline{$1$}}&$0.9867$&$0.9767$&$0.9867$&$0.9400$&$0.9933$&\green{\underline{$1$}}$0.8600$&$0.4067$&$0.4467$&$0.4033$&$0.8467$&$0.9867$&$0.8533$\\
Trace&\green{\underline{$1$}}&\green{\underline{$1$}}&\green{\underline{$1$}}&\green{\underline{$1$}}&\green{\underline{$1$}}&\green{\underline{$1$}}&\green{\underline{$1$}}&\green{\underline{$1$}}&\green{\underline{$1$}}&\green{\underline{$1$}}&\green{\underline{$1$}}&\green{\underline{$1$}}&$0.9900$&\green{\underline{$1$}}&\green{\underline{$1$}}\\
\end{tabular}
}
\label{results:ucrmedians}
\end{table*}

\begin{figure*}[!h]
\centering

\begin{tabular}{lr}

\begingroup
\setlength{\tabcolsep}{2pt} 
\begin{tabular}{ccc}

(a)  & (b) & (c) \\ 
\includegraphics[width=0.14\linewidth, trim={3mm 3mm 3mm 3mm},clip]{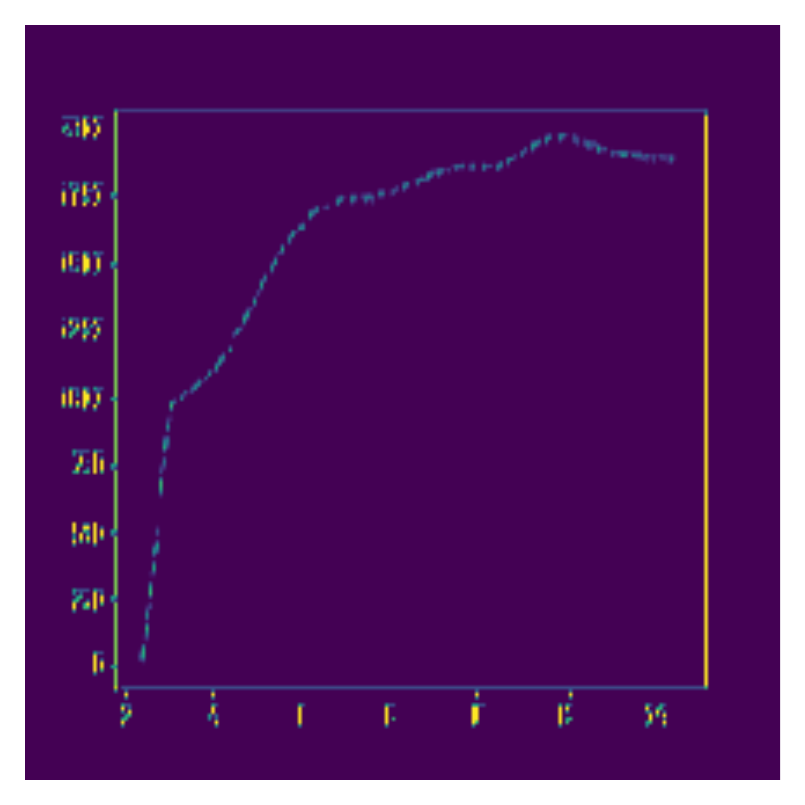} & 
\includegraphics[width=0.14\linewidth, trim={3mm 3mm 3mm 3mm},clip]{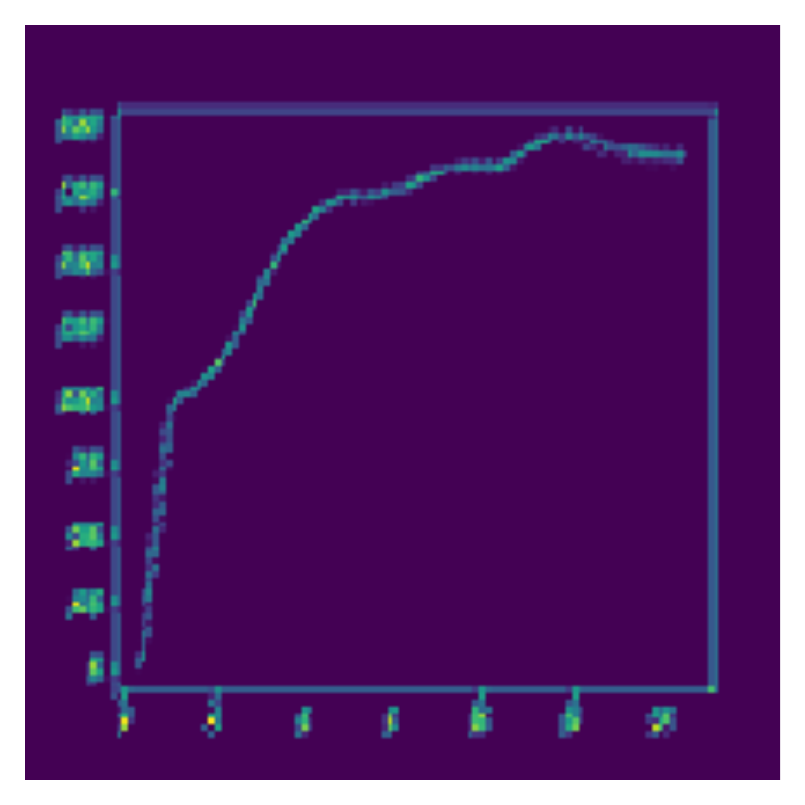} & 
\includegraphics[width=0.14\linewidth, trim={3mm 3mm 3mm 3mm},clip]{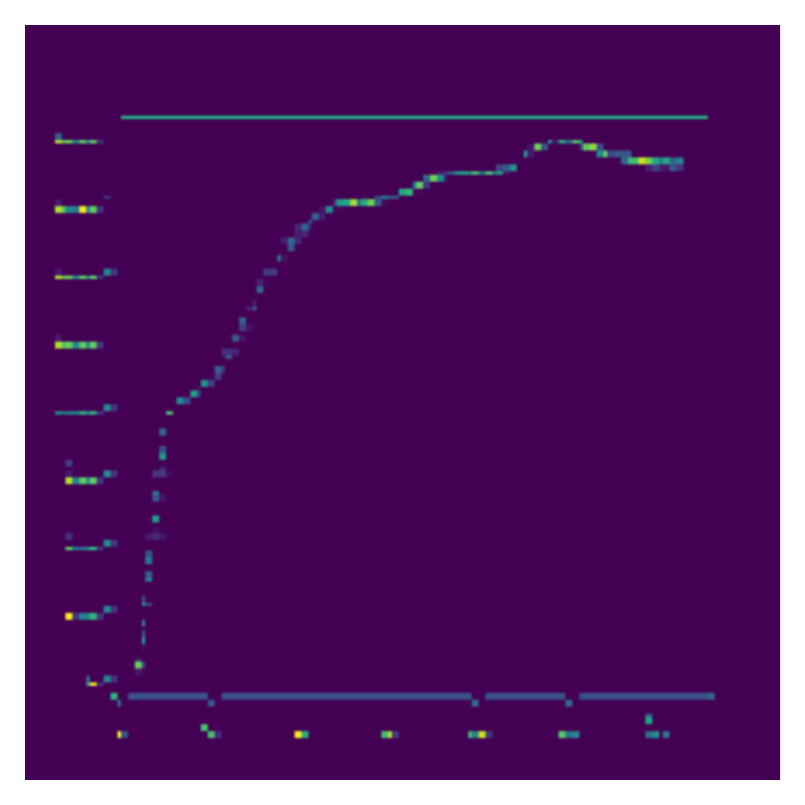}
\\
(d)  & (e)  & (f) \\
\includegraphics[width=0.14\linewidth, trim={3mm 3mm 3mm 3mm},clip]{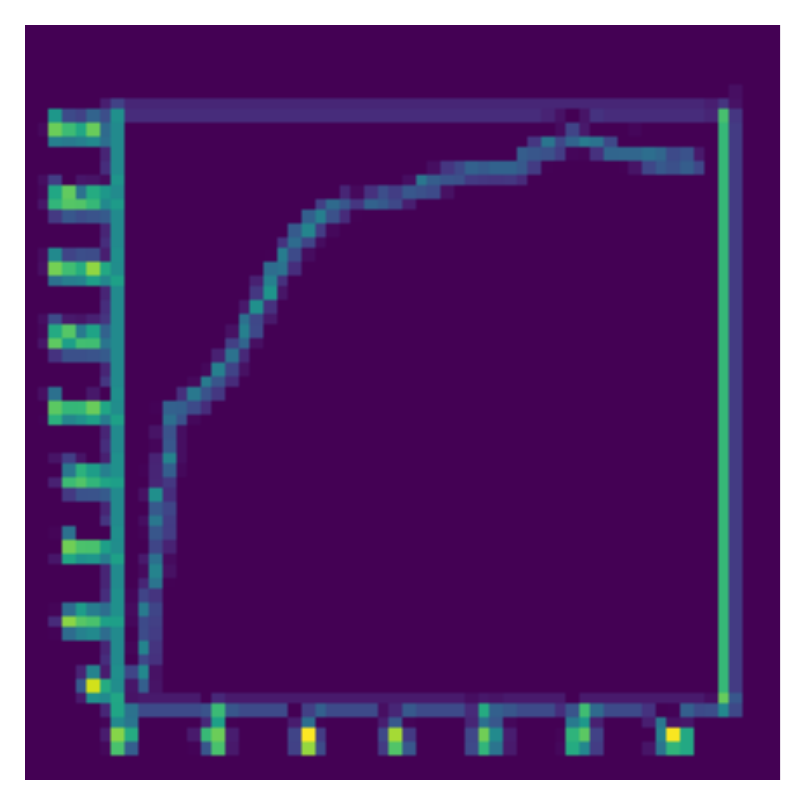} & 
\includegraphics[width=0.14\linewidth, trim={3mm 3mm 3mm 3mm},clip]{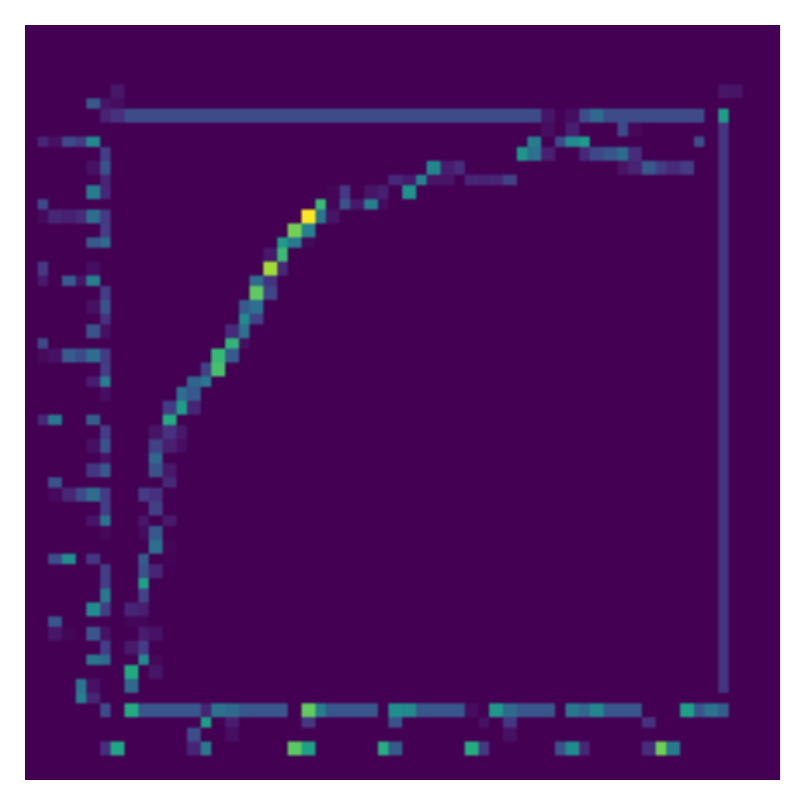} & 
\includegraphics[width=0.14\linewidth, trim={3mm 3mm 3mm 3mm},clip]{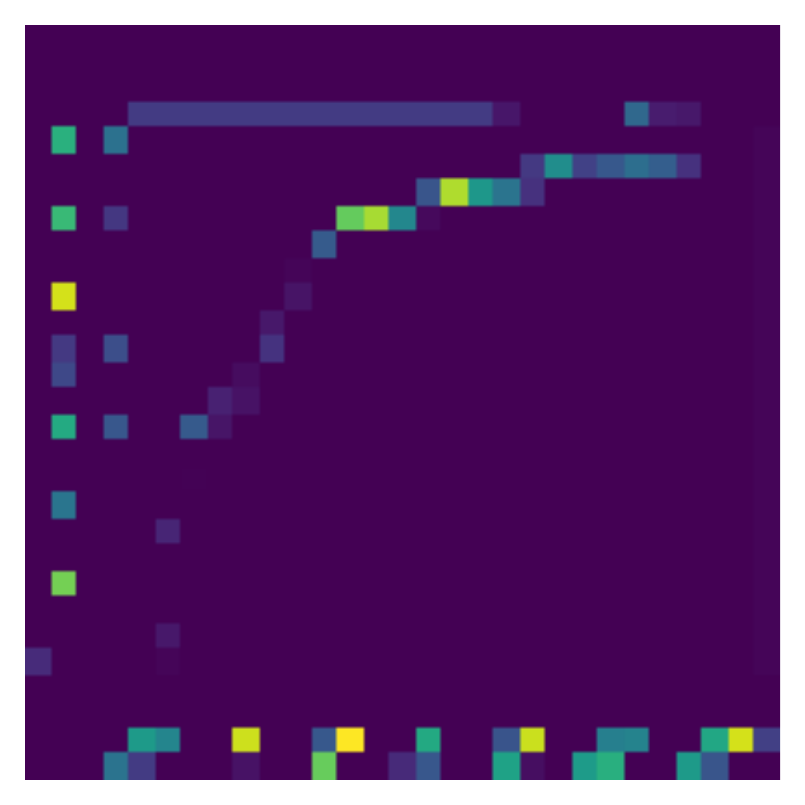}

\end{tabular}
\endgroup
 
 &
 
\begingroup
\setlength{\tabcolsep}{2pt} 
\begin{tabular}{ccc}

(a)  & (b)  & (c) \\ 
\includegraphics[width=0.14\linewidth, trim={3mm 3mm 3mm 3mm},clip]{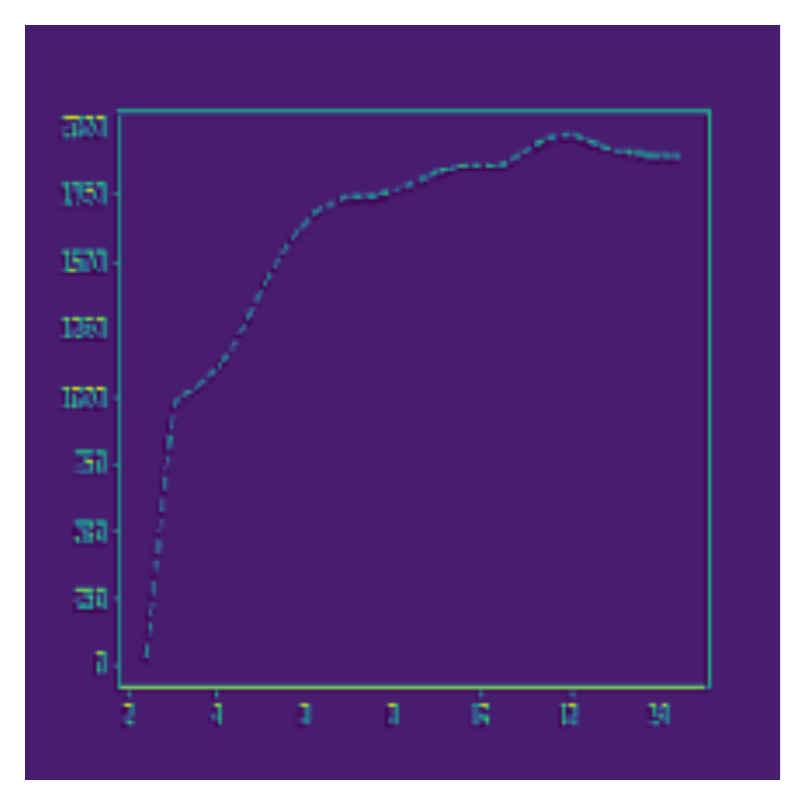} & 
\includegraphics[width=0.14\linewidth, trim={3mm 3mm 3mm 3mm},clip]{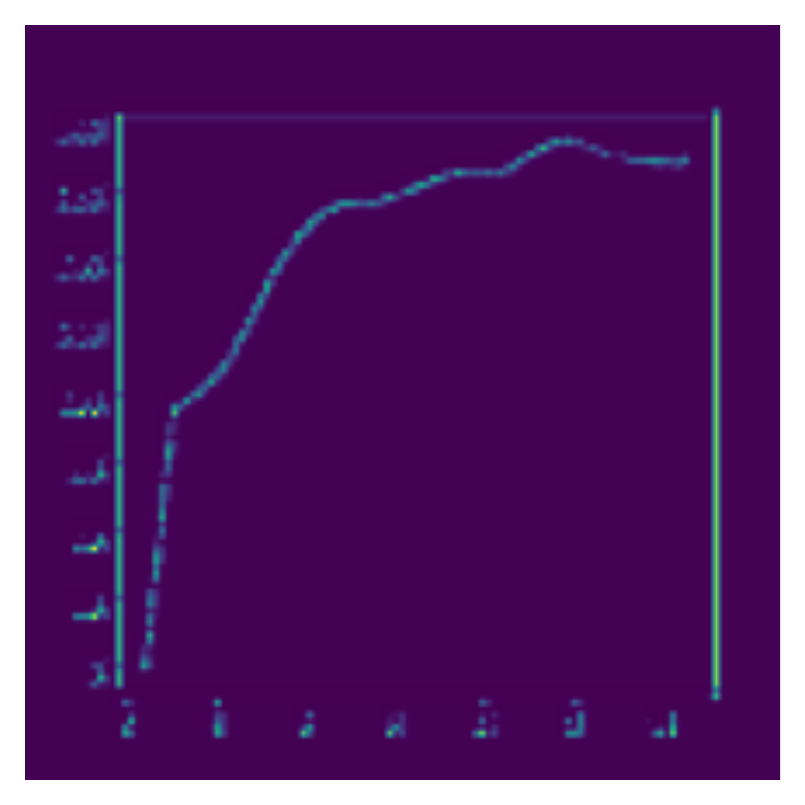} & 
\includegraphics[width=0.14\linewidth, trim={3mm 3mm 3mm 3mm},clip]{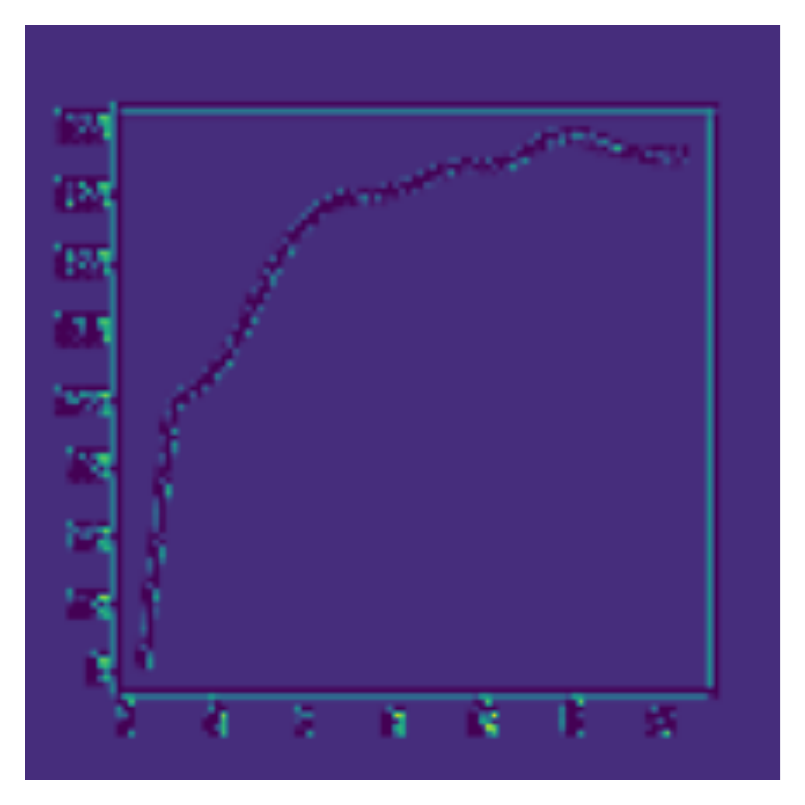}
\\
(d)  & (e)  & (f) \\
\includegraphics[width=0.14\linewidth, trim={3mm 3mm 3mm 3mm},clip]{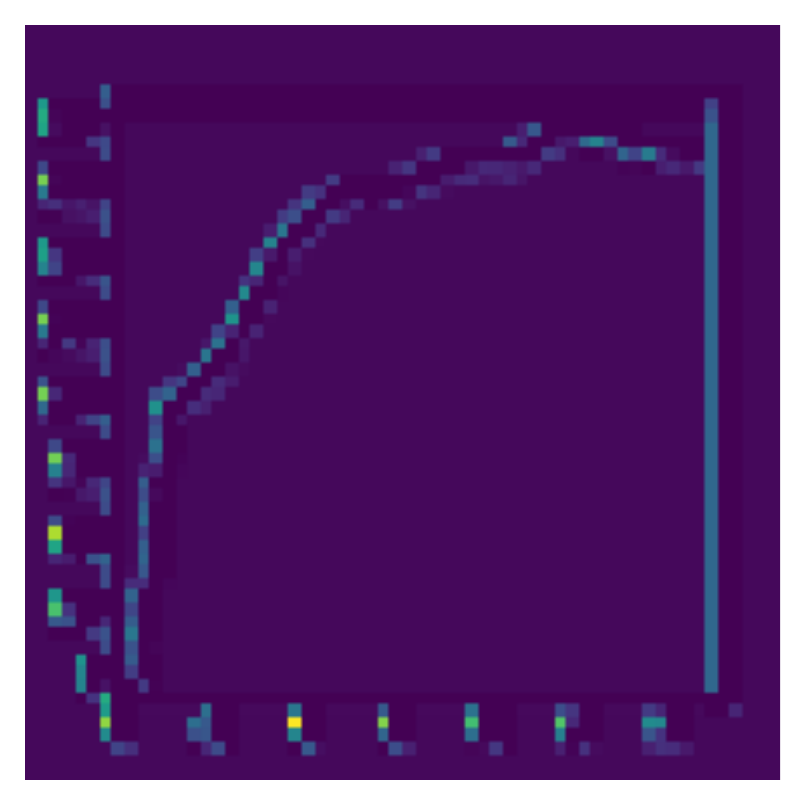} & 
\includegraphics[width=0.14\linewidth, trim={3mm 3mm 3mm 3mm},clip]{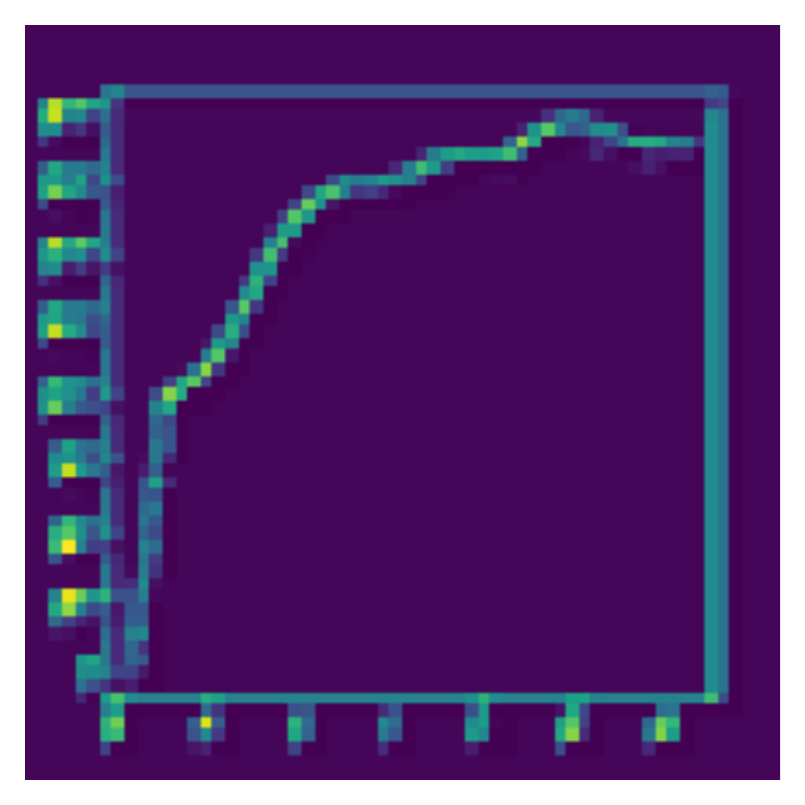} & 
\includegraphics[width=0.14\linewidth, trim={3mm 3mm 3mm 3mm},clip]{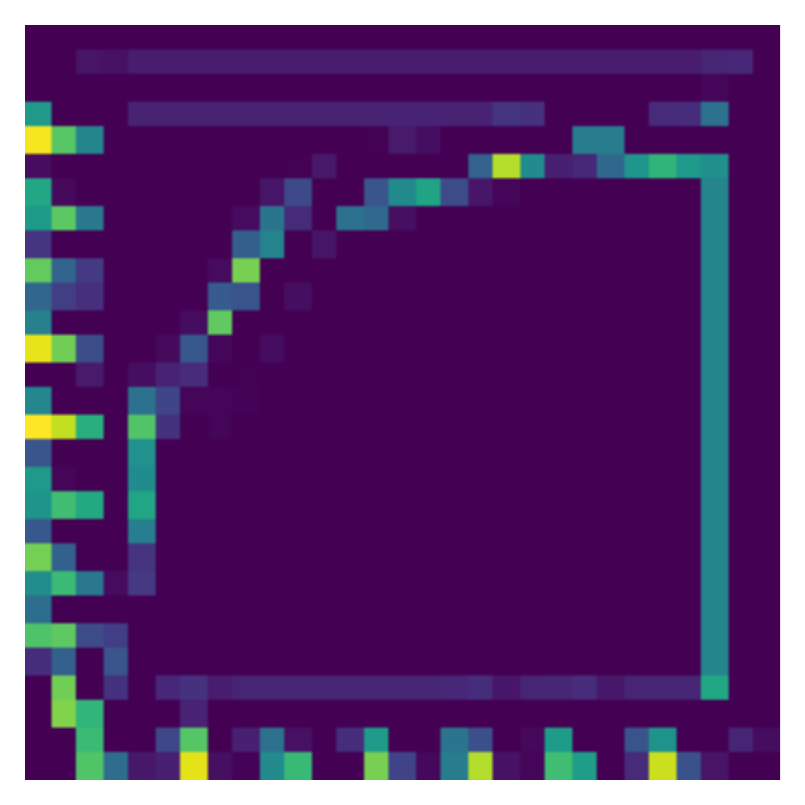}

\end{tabular}
\endgroup

\end{tabular}

\caption{Feature maps from different convolutional layers taken from two models trained on the OPTOX dataset. 
The set of images on the left side was obtained using input that was normalized, while the right side corresponds to non normalized images.
For both set of plots a) was taken from the first convolution layer, b) and c) from the second, d) and e) from the third, and f) from the fourth.}
\label{figure:fmaps}

\end{figure*}

\subsection*{The effect of normalizing image data}
\label{subsection:contribution_normalization}

Before deciding to apply the samplewise normalization described in Section~\ref{subsection:normalizing_image_data}, we performed some trial runs where the only pre-processing of image data was the rescaling of the pixel values. The main difference we observed was that, without normalization, there were multiple runs that converged to sub-optimal solutions during training, particularly on the UCR problems. Furthermore, even when the models converged to good solutions, they took much longer to converge than the ones with normalization, as much as $4 \times$ longer on the OPTOX dataset.

A possible explanation for this difference is that, without normalization, the filters have to learn the features from low pixel values (curves and axes) that are surrounded by high pixel values (white space), a much more difficult task than the opposite, that is to learn features from the pixels that already enter the network with the highest values. Indeed, when analysing feature maps from models trained with non normalized image data, we observed that sometimes the time series are not even detected in the first convolutional layer.
On later layers there are in fact feature maps showing that the filters identified key information such as the time series and axes. However, there are a large number of feature maps where no prominent features are identified, something that does not happen when the images are normalized.

Figure~\ref{figure:fmaps} presents two sets of feature maps. On the left, feature maps obtained from a model trained with normalized data; on the right, feature maps from a model trained on the same dataset (and same training/test partition) without normalization. Comparing both sets, we can see that, with normalization, the filters begin to detect different important areas of the image, starting on the first convolutional layer, whereas without normalization there are almost no activations on the feature maps of the first and second convolutional layers.


\section{Conclusions and Future Work}

We have proposed a new approach for time series classification where the time series are represented as plot images and given to a shallow CNN. Our methodology is very simple and can be applied to a broad variety of problems, including ones where different time series have different lengths. We tested our methodology on two real-world non public datasets, where it outperformed all other methods, and on the UCR archive, where it was able to beat the current state-of-the-art methods in a small set of problems. We have also shown that, thanks to the image pre-processing we apply, the models are able to converge faster and detect important image features earlier during training.
We conclude that, if a simple naive design like ours can obtain such good results, then there is much to explore when using deep learning methods that rely on image data.

As future research, we will test our methodology using known network architectures such as ResNet, VGG or Inception, and alternative image representations of the data. Finally, we will use the same approach to solve other classification problems that may not involve time series, but rather signal data, such as remote sensing applications.

\section*{Acknowledgements}
This work was partially supported by FCT through funding of Research Units LASIGE (UIDB/ 00408/2020 and UIDP/ 00408/2020) and MARE (UIDB/ 04292/2020); \mbox{AICE} (DSAIPA/DS/ 0113/2019), Projects \mbox{BINDER} (PTDC/CCI-INF/ 29168/2017), \mbox{GADgET} (DSAIPA/DS/ 0022/2018), \mbox{INTERPHENO} (PTDC/ASP-PLA/ 28726/2017), \mbox{OPTOX} (PTDC/CTA-AMB/ 30056/2017), \mbox{PREDICT} (PTDC/CCI-CIF/ 29877/2017); PhD Grant(SFRH/BD/143972/2019); Research Contracts CEECIND/00511/2017 and CEECIND/02513/2017. The authors also acknowledge Prof. Alberto Rainoldi and Dr. Marco Ivaldi for the useful discussion on the physiological dataset and the Centre of Preventive Medicine and Sport - SUISM - University Structure of Hygiene and Sport Sciences, Centre of Excellence of the University of Torino for allowing the use of the dataset.



\bibliographystyle{named}
\bibliography{ijcai21-multiauthor}

\begin{thebibliography}{}

\bibitem[\protect\citeauthoryear{Azzali \bgroup \em et al.\egroup
  }{2020}]{Azzali2020TowardsTU}
Irene Azzali, L.~Vanneschi, Illya Bakurov, Sara Silva, M.~Ivaldi, and
  M.~Giacobini.
\newblock Towards the use of vector based gp to predict physiological time
  series.
\newblock {\em Appl. Soft Comput.}, 89:106097, 2020.

\bibitem[\protect\citeauthoryear{Breiman}{2001}]{Breiman2001}
Leo Breiman.
\newblock Random forests.
\newblock {\em Machine Learning}, 45:5--32, 2001.

\bibitem[\protect\citeauthoryear{Chen and Guestrin}{2016}]{2016XGBoostAS}
T.~Chen and Carlos Guestrin.
\newblock Xgboost: A scalable tree boosting system.
\newblock {\em Proceedings of the 22nd ACM SIGKDD International Conference on
  Knowledge Discovery and Data Mining}, 2016.

\bibitem[\protect\citeauthoryear{Dau \bgroup \em et al.\egroup
  }{2018}]{UCRArchive2018}
Hoang~Anh Dau, Eamonn Keogh, Kaveh Kamgar, Chin-Chia~Michael Yeh, Yan Zhu,
  Shaghayegh Gharghabi, Chotirat~Ann Ratanamahatana, Yanping, Bing Hu, Nurjahan
  Begum, Anthony Bagnall, Abdullah Mueen, Gustavo Batista, and Hexagon-ML.
\newblock The ucr time series classification archive, October 2018.
\newblock \url{https://www.cs.ucr.edu/~eamonn/time_series_data_2018/}.

\bibitem[\protect\citeauthoryear{Dempster \bgroup \em et al.\egroup
  }{2020}]{Dempster2020ROCKETEF}
Angus Dempster, Franccois Petitjean, and Geoffrey~I. Webb.
\newblock Rocket: exceptionally fast and accurate time series classification
  using random convolutional kernels.
\newblock {\em Data Mining and Knowledge Discovery}, 34:1454--1495, 2020.

\bibitem[\protect\citeauthoryear{Dovhalets \bgroup \em et al.\egroup
  }{2018}]{L4}
Dmytro Dovhalets, Boris Kovalerchuk, Szil{\'{a}}rd Vajda, and R{\u{a}}zvan
  Andonie.
\newblock Deep learning of 2-d images representing n-d data in general line
  coordinates.
\newblock In {\em International Symposium on Affective Science and
  Engineering}, volume {ISASE}2018, pages 1--6. Japan Society of Kansei
  Engineering, 2018.

\bibitem[\protect\citeauthoryear{Fawaz \bgroup \em et al.\egroup
  }{2020}]{Fawaz2020InceptionTimeFA}
Hassan~Ismail Fawaz, B.~Lucas, G.~Forestier, Charlotte Pelletier, D.~Schmidt,
  Jonathan Weber, Geoffrey~I. Webb, L.~Idoumghar, Pierre-Alain Muller, and
  Franccois Petitjean.
\newblock Inceptiontime: Finding alexnet for time series classification.
\newblock {\em ArXiv}, abs/1909.04939, 2020.

\bibitem[\protect\citeauthoryear{Hunter}{2007}]{Hunter:2007}
J.~D. Hunter.
\newblock Matplotlib: A 2d graphics environment.
\newblock {\em Computing in Science \& Engineering}, 9(3):90--95, 2007.

\bibitem[\protect\citeauthoryear{{Karimi-Bidhendi} \bgroup \em et al.\egroup
  }{2018}]{L7}
S.~{Karimi-Bidhendi}, F.~{Munshi}, and A.~{Munshi}.
\newblock Scalable classification of univariate and multivariate time series.
\newblock In {\em 2018 IEEE International Conference on Big Data}, pages
  1598--1605, 2018.

\bibitem[\protect\citeauthoryear{{Keim}}{2000}]{L1}
D.~A. {Keim}.
\newblock Designing pixel-oriented visualization techniques: theory and
  applications.
\newblock {\em IEEE Transactions on Visualization and Computer Graphics},
  6(1):59--78, 2000.

\bibitem[\protect\citeauthoryear{Kovalerchuk}{2018}]{L3}
Boris Kovalerchuk.
\newblock {\em Visual Knowledge Discovery and Machine Learning}.
\newblock Springer International Publishing, 2018.

\bibitem[\protect\citeauthoryear{Large \bgroup \em et al.\egroup
  }{2018}]{Large2018FromBT}
J.~Large, Anthony~J. Bagnall, S.~Malinowski, and R.~Tavenard.
\newblock From bop to boss and beyond: Time series classification with
  dictionary based classifiers.
\newblock {\em ArXiv}, abs/1809.06751, 2018.

\bibitem[\protect\citeauthoryear{Lines \bgroup \em et al.\egroup
  }{2016}]{Lines2016HIVECOTETH}
J.~Lines, Sarah Taylor, and Anthony~J. Bagnall.
\newblock Hive-cote: The hierarchical vote collective of transformation-based
  ensembles for time series classification.
\newblock {\em 2016 IEEE ICDM}, pages 1041--1046, 2016.

\bibitem[\protect\citeauthoryear{Lucas \bgroup \em et al.\egroup
  }{2019}]{Lucas2019ProximityFA}
B.~Lucas, Ahmed Shifaz, Charlotte Pelletier, Lachlan O'Neill, N.~A. Zaidi,
  B.~Goethals, François Petitjean, and Geoffrey~I. Webb.
\newblock Proximity forest: an effective and scalable distance-based classifier
  for time series.
\newblock {\em Data Mining and Knowledge Discovery}, 33:607--635, 2019.

\bibitem[\protect\citeauthoryear{Lyu and Haque}{2018}]{L5}
Boyu Lyu and Anamul Haque.
\newblock Deep learning based tumor type classification using gene expression
  data.
\newblock {\em bioRxiv}, 2018.

\bibitem[\protect\citeauthoryear{Sch{\"a}fer and
  Leser}{2017}]{Schfer2017FastAA}
Patrick Sch{\"a}fer and U.~Leser.
\newblock Fast and accurate time series classification with weasel.
\newblock {\em Proceedings of the 2017 ACM on Conference on Information and
  Knowledge Management}, 2017.

\bibitem[\protect\citeauthoryear{Sch{\"a}fer}{2014}]{Schfer2014TheBI}
Patrick Sch{\"a}fer.
\newblock The boss is concerned with time series classification in the presence
  of noise.
\newblock {\em Data Mining and Knowledge Discovery}, 29:1505--1530, 2014.

\bibitem[\protect\citeauthoryear{Sharma \bgroup \em et al.\egroup }{2019}]{L6}
Alok Sharma, Edwin Vans, Daichi Shigemizu, Keith~A. Boroevich, and Tatsuhiko
  Tsunoda.
\newblock {DeepInsight}: A methodology to transform a non-image data to an
  image for convolution neural network architecture.
\newblock {\em Scientific Reports}, 9(1), August 2019.

\bibitem[\protect\citeauthoryear{Shifaz \bgroup \em et al.\egroup
  }{2020}]{Shifaz2020TSCHIEFAS}
Ahmed Shifaz, Charlotte Pelletier, F.~Petitjean, and Geoffrey~I. Webb.
\newblock Ts-chief: a scalable and accurate forest algorithm for time series
  classification.
\newblock {\em Data Mining and Knowledge Discovery}, 34:742--775, 2020.

\bibitem[\protect\citeauthoryear{Silva \bgroup \em et al.\egroup
  }{2020}]{Silva2020ComfortablyNE}
M.~Silva, Eduardo Feij{\~a}o, Ricardo da~Cruz~de Carvalho, Irina~A. Duarte,
  A.~Matos, M.~T. Cabrita, A.~Barreiro, M.~Lemos, S.~Novais, Jo{\~a}o~H
  Marques, I.~Caçador, P.~Reis-Santos, V.~Fonseca, and B.~Duarte.
\newblock Comfortably numb: Ecotoxicity of the non-steroidal anti-inflammatory
  drug ibuprofen on phaeodactylum tricornutum.
\newblock {\em Marine environmental research}, 161:105109, 2020.

\bibitem[\protect\citeauthoryear{van~der Maaten and Hinton}{2008}]{L2}
Laurens van~der Maaten and Geoffrey Hinton.
\newblock Visualizing data using t-sne.
\newblock {\em Journal of Machine Learning Research}, 9(86):2579--2605, 2008.

\end{thebibliography}


@ARTICLE{L1,
  author={D. A. {Keim}},
  journal={IEEE Transactions on Visualization and Computer Graphics},
  title={Designing pixel-oriented visualization techniques: theory and applications},
  year={2000},
  volume={6},
  number={1},
  pages={59-78}}
 
 @article{L2,
  author  = {Laurens van der Maaten and Geoffrey Hinton},
  title   = {Visualizing Data using t-SNE},
  journal = {Journal of Machine Learning Research},
  year    = {2008},
  volume  = {9},
  number  = {86},
  pages   = {2579-2605}
}

@book{L3,
  year = {2018},
  publisher = {Springer International Publishing},
  author = {Boris Kovalerchuk},
  title = {Visual Knowledge Discovery and Machine Learning}
}

@INPROCEEDINGS{L4,
  year = {2018},
  publisher = {Japan Society of Kansei Engineering},
  volume = {{ISASE}2018},
  number = {0},
  pages = {1--6},
  author = {Dmytro Dovhalets and Boris Kovalerchuk and Szil{\'{a}}rd Vajda and R{\u{a}}zvan Andonie},
  title = {Deep Learning of 2-D Images Representing n-D Data in General Line Coordinates},
  booktitle = {International Symposium on Affective Science and Engineering}
}

@article {L5,
author = {Lyu, Boyu and Haque, Anamul},
title = {Deep Learning Based Tumor Type Classification Using Gene Expression Data},
elocation-id = {364323},
year = {2018},
doi = {10.1101/364323},
publisher = {Cold Spring Harbor Laboratory},
journal = {bioRxiv}
}

@article{L6,
  year = {2019},
  month = aug,
  publisher = {Springer Science and Business Media {LLC}},
  volume = {9},
  number = {1},
  author = {Alok Sharma and Edwin Vans and Daichi Shigemizu and Keith A. Boroevich and Tatsuhiko Tsunoda},
  title = {{DeepInsight}: A methodology to transform a non-image data to an image for convolution neural network architecture},
  journal = {Scientific Reports}
}

@INPROCEEDINGS{L7,
  author={S. {Karimi-Bidhendi} and F. {Munshi} and A. {Munshi}},
  booktitle={2018 IEEE International Conference on Big Data},
  title={Scalable Classification of Univariate and Multivariate Time Series},
  year={2018},
  volume={},
  number={},
  pages={1598-1605},}

@article{Fawaz2020InceptionTimeFA,
  title={InceptionTime: Finding AlexNet for Time Series Classification},
  author={Hassan Ismail Fawaz and B. Lucas and G. Forestier and Charlotte Pelletier and D. Schmidt and Jonathan Weber and Geoffrey I. Webb and L. Idoumghar and Pierre-Alain Muller and Franccois Petitjean},
  journal={ArXiv},
  year={2020},
  volume={abs/1909.04939}
}



@article{Chawla2002SMOTESM,
  title={SMOTE: Synthetic Minority Over-sampling Technique},
  author={Nitesh V. Chawla and K. Bowyer and L. Hall and W. P. Kegelmeyer},
  journal={J. Artif. Intell. Res.},
  year={2002},
  volume={16},
  pages={321-357}
}

@misc{UCRArchive2018,
title = {The UCR Time Series Classification Archive},
author = {Dau, Hoang Anh and Keogh, Eamonn and Kamgar, Kaveh and Yeh, Chin-Chia Michael and Zhu, Yan 
          and Gharghabi, Shaghayegh and Ratanamahatana, Chotirat Ann and Yanping and Hu, Bing 
          and Begum, Nurjahan and Bagnall, Anthony and Mueen, Abdullah and Batista, Gustavo and Hexagon-ML},
year = {2018},
month = {October},
note = {\url{https://www.cs.ucr.edu/~eamonn/time_series_data_2018/}}
}


@article{KarimiBidhendi2018ScalableCO,
  title={Scalable Classification of Univariate and Multivariate Time Series},
  author={Saeed Karimi-Bidhendi and Faramarz Munshi and Ashfaq Munshi},
  journal={2018 IEEE International Conference on Big Data},
  year={2018},
  pages={1598-1605}
}

@article{GLC,
author = {Dovhalets, Dmytro and Kovalerchuk, Boris and Vajda, Szilard and Andonie, Razvan},
year = {2018},
month = {11},
pages = {1-6},
title = {Deep Learning of 2-D Images Representing n-D Data in General Line Coordinates},
volume = {ISASE2018},
journal = {International Symposium on Affective Science and Engineering},
doi = {10.5057/isase.2018-C000025}
}

@article{Lyu2018DeepLB,
  title={Deep Learning Based Tumor Type Classification Using Gene Expression Data},
  author={Boyu Lyu and A. Haque},
  journal={Proceedings of the 2018 ACM International Conference on Bioinformatics, Computational Biology, and Health Informatics},
  year={2018}
}

@article{Schfer2014TheBI,
  title={The BOSS is concerned with time series classification in the presence of noise},
  author={Patrick Sch{\"a}fer},
  journal={Data Mining and Knowledge Discovery},
  year={2014},
  volume={29},
  pages={1505-1530}
}

@article{Large2018FromBT,
  title={From BOP to BOSS and Beyond: Time Series Classification with Dictionary Based Classifiers},
  author={J. Large and Anthony J. Bagnall and S. Malinowski and R. Tavenard},
  journal={ArXiv},
  year={2018},
  volume={abs/1809.06751}
}

@article{Lines2016HIVECOTETH,
  title={HIVE-COTE: The Hierarchical Vote Collective of Transformation-Based Ensembles for Time Series Classification},
  author={J. Lines and Sarah Taylor and Anthony J. Bagnall},
  journal={2016 IEEE ICDM},
  year={2016},
  pages={1041-1046}
}

@article{Dempster2020ROCKETEF,
  title={ROCKET: exceptionally fast and accurate time series classification using random convolutional kernels},
  author={Angus Dempster and Franccois Petitjean and Geoffrey I. Webb},
  journal={Data Mining and Knowledge Discovery},
  year={2020},
  volume={34},
  pages={1454-1495}
}

@article{Lucas2019ProximityFA,
  title={Proximity Forest: an effective and scalable distance-based classifier for time series},
  author={B. Lucas and Ahmed Shifaz and Charlotte Pelletier and Lachlan O'Neill and N. A. Zaidi and B. Goethals and François Petitjean and Geoffrey I. Webb},
  journal={Data Mining and Knowledge Discovery},
  year={2019},
  volume={33},
  pages={607-635}
}

@article{Shifaz2020TSCHIEFAS,
  title={TS-CHIEF: a scalable and accurate forest algorithm for time series classification},
  author={Ahmed Shifaz and Charlotte Pelletier and F. Petitjean and Geoffrey I. Webb},
  journal={Data Mining and Knowledge Discovery},
  year={2020},
  volume={34},
  pages={742-775}
}

@article{Schfer2017FastAA,
  title={Fast and Accurate Time Series Classification with WEASEL},
  author={Patrick Sch{\"a}fer and U. Leser},
  journal={Proceedings of the 2017 ACM on Conference on Information and Knowledge Management},
  year={2017}
}

@article{Silva2020ComfortablyNE,
  title={Comfortably numb: Ecotoxicity of the non-steroidal anti-inflammatory drug ibuprofen on Phaeodactylum tricornutum.},
  author={M. Silva and Eduardo Feij{\~a}o and Ricardo da Cruz de Carvalho and Irina A. Duarte and A. Matos and M. T. Cabrita and A. Barreiro and M. Lemos and S. Novais and Jo{\~a}o H Marques and I. Caçador and P. Reis-Santos and V. Fonseca and B. Duarte},
  journal={Marine environmental research},
  year={2020},
  volume={161},
  pages={
          105109
        }
}

@article{Azzali2020TowardsTU,
  title={Towards the use of vector based GP to predict physiological time series},
  author={Irene Azzali and L. Vanneschi and Illya Bakurov and Sara Silva and M. Ivaldi and M. Giacobini},
  journal={Appl. Soft Comput.},
  year={2020},
  volume={89},
  pages={106097}
}

@Article{Hunter:2007,
  Author    = {Hunter, J. D.},
  Title     = {Matplotlib: A 2D graphics environment},
  Journal   = {Computing in Science \& Engineering},
  Volume    = {9},
  Number    = {3},
  Pages     = {90--95},
  abstract  = {Matplotlib is a 2D graphics package used for Python for
  application development, interactive scripting, and publication-quality
  image generation across user interfaces and operating systems.},
  publisher = {IEEE COMPUTER SOC},
  doi       = {10.1109/MCSE.2007.55},
  year      = 2007
}



@article{2016XGBoostAS,
  title={XGBoost: A Scalable Tree Boosting System},
  author={T. Chen and Carlos Guestrin},
  journal={Proceedings of the 22nd ACM SIGKDD International Conference on Knowledge Discovery and Data Mining},
  year={2016}
}

@Article{Breiman2001,
author="Breiman, Leo",
title="Random Forests",
journal="Machine Learning",
year="2001",
volume="45",
pages="5--32"
}


\clearpage

\appendix
\section*{Appendix}
\label{appendix}
\section{OPTOX $p$-values}
\label{sections:OPTOX-pvals}

\begin{table}[!h]
\centering
\caption{Kruskal-Wallis \textit{p}-values comparing all methods of the
contaminant prediction task of the OPTOX dataset. 
Above the diagonal, we have the results for the test set, and below for the training set. Significant differences ($p<0.01$) are represented in \green{\underline{green}} or \red{\textit{red}} when the method on the left is significantly better or worse, respectively.}
\resizebox{\linewidth}{!}{
\begin{tabular}{c|cccccccc}
&RF&XG&ROCKET&ANN&1D CNN&CNN Log&CNN&\\
\hline
RF&--------&\red{$\mathit{1.23e^{-10}}$}&\red{$\mathit{2.58e^{-11}}$}&\red{$\mathit{2.80e^{-11}}$}&\red{$\mathit{1.89e^{-08}}$}&\red{$\mathit{2.68e^{-11}}$}&\red{$\mathit{2.63e^{-11}}$}\parbox[c]{0mm}{\multirow{7}{*}{\rotatebox[origin=t]{90}{\makecell{\\Test}}}}\\
XG&\green{\underline{$2.24e^{-11}$}}&--------&\red{$\mathit{2.60e^{-11}}$}&\red{$\mathit{2.82e^{-11}}$}&\red{$\mathit{1.01e^{-07}}$}&\red{$\mathit{2.70e^{-11}}$}&\red{$\mathit{2.65e^{-11}}$}\\
ROCKET&\green{\underline{$1.88e^{-11}$}}&\green{\underline{$1.08e^{-05}$}}&--------&\green{\underline{$2.59e^{-11}$}}&\green{\underline{$4.04e^{-03}$}}&\red{$\mathit{5.79e^{-04}}$}&$3.64e^{-02}$\\
ANN&\red{$\mathit{2.82e^{-11}}$}&\red{$\mathit{2.26e^{-11}}$}&\red{$\mathit{1.90e^{-11}}$}&--------&\red{$\mathit{7.04e^{-04}}$}&\red{$\mathit{2.69e^{-11}}$}&\red{$\mathit{2.64e^{-11}}$}\\
1D CNN&\green{\underline{$2.34e^{-11}$}}&$8.89e^{-02}$&$8.08e^{-01}$&\green{\underline{$2.36e^{-11}$}}&--------&\red{$\mathit{2.68e^{-06}}$}&\red{$\mathit{3.19e^{-04}}$}\\
CNN Log&\green{\underline{$1.63e^{-09}$}}&\red{$\mathit{2.07e^{-11}}$}&\red{$\mathit{1.73e^{-11}}$}&\green{\underline{$2.61e^{-11}$}}&\red{$\mathit{6.55e^{-09}}$}&--------&\green{\underline{$6.93e^{-03}$}}\\
CNN&\green{\underline{$2.54e^{-11}$}}&\red{$\mathit{2.03e^{-11}}$}&\red{$\mathit{1.70e^{-11}}$}&\green{\underline{$2.56e^{-11}$}}&\red{$\mathit{3.09e^{-08}}$}&$2.55e^{-02}$&--------\\
&\multicolumn{7}{c}{Training}\\
\end{tabular}
}

\label{pvalues_xeno}
\end{table}

\section{FISIO $p$-values}
\label{sections:Fisio-pvals}


\begin{table}[!h]
\centering
\caption{Kruskal-Wallis \textit{p}-values comparing all methods of the BMI\_Siamese prediction task. Above the diagonal, we have the results for the test set, and below for the training set. Significant results~($p < 0.01$) are represented in \green{\underline{green}} or \red{\textit{red}} when the method on the left is significantly better or worse, respectively.}
\resizebox{0.75\linewidth}{!}{
\begin{tabular}{c|cccc}
BMI\_Siamese&ANN&1D CNN&CNN&\\
\hline
ANN&--------&\green{\underline{$8.31e^{-10}$}}&$4.76e^{-01}$\parbox[c]{0mm}{\multirow{4}{*}{\rotatebox[origin=t]{90}{\makecell{\\Test}}}}\\
1D CNN&\green{\underline{$1.12e^{-11}$}}&--------&\red{$\mathit{1.32e^{-09}}$}\\
CNN&\green{\underline{$1.04e^{-12}$}}&\green{\underline{$6.33e^{-04}$}}&--------\\
&\multicolumn{3}{c}{Training}\\
\end{tabular}
}
\label{pvalues_BMI_Siamese}
\end{table}


\begin{table}[!h]
\centering
\caption{Kruskal-Wallis \textit{p}-values comparing all methods of the BMI\_HR prediction task. Above the diagonal, we have the results for the test set, and below for the training set. Significant results~($p < 0.01$) are represented in \green{\underline{green}} or \red{\textit{red}} when the method on the left is significantly better or worse, respectively.}
\resizebox{\linewidth}{!}{\begin{tabular}{c|ccccccc}
BMI\_HR&RF&XG&Rocket&ANN&1D CNN&CNN&\\
\hline
RF&--------&$9.94e^{-01}$&\green{\underline{$2.34e^{-08}$}}&\green{\underline{$7.81e^{-07}$}}&\green{\underline{$2.25e^{-11}$}}&$6.55e^{-01}$\parbox[c]{0mm}{\multirow{7}{*}{\rotatebox[origin=t]{90}{\makecell{\\Test}}}}\\
XG&$1.00e^{+00}$&--------&\green{\underline{$3.11e^{-08}$}}&\green{\underline{$2.96e^{-06}$}}&\green{\underline{$2.24e^{-11}$}}&$7.32e^{-01}$\\
Rocket&\green{\underline{$1.10e^{-12}$}}&\green{\underline{$1.12e^{-12}$}}&--------&\red{$\mathit{3.52e^{-04}}$}&\green{\underline{$7.29e^{-05}$}}&\red{$\mathit{1.48e^{-08}}$}\\
ANN&\green{\underline{$5.89e^{-08}$}}&\green{\underline{$1.30e^{-10}$}}&\red{$\mathit{1.06e^{-12}}$}&--------&\green{\underline{$1.22e^{-10}$}}&\red{$\mathit{1.86e^{-08}}$}\\
1D CNN&\green{\underline{$8.88e^{-12}$}}&\green{\underline{$7.31e^{-12}$}}&\red{$\mathit{5.35e^{-03}}$}&\green{\underline{$8.18e^{-12}$}}&--------&\red{$\mathit{1.96e^{-11}}$}\\
CNN&\green{\underline{$1.10e^{-12}$}}&\green{\underline{$1.12e^{-12}$}}&$1.00e^{+00}$&\green{\underline{$1.06e^{-12}$}}&\green{\underline{$5.35e^{-03}$}}&--------\\
&\multicolumn{6}{c}{Training}\\
\end{tabular}
}\label{pvalues_BMI_HR}
\end{table}


\begin{table}[!h]
\centering
\caption{Kruskal-Wallis \textit{p}-values comparing all methods of the BMI\_VEN prediction task. Above the diagonal, we have the results for the test set, and below for the training set. Significant results~($p < 0.01$) are represented in \green{\underline{green}} or \red{\textit{red}} when the method on the left is significantly better or worse, respectively.}
\resizebox{\linewidth}{!}{\begin{tabular}{c|ccccccc}
BMI\_VEN&RF&XG&Rocket&ANN&1D CNN&CNN&\\
\hline
RF&--------&$9.11e^{-01}$&\green{\underline{$2.33e^{-07}$}}&\green{\underline{$4.21e^{-03}$}}&\green{\underline{$8.82e^{-11}$}}&$3.97e^{-01}$\parbox[c]{0mm}{\multirow{7}{*}{\rotatebox[origin=t]{90}{\makecell{\\Test}}}}\\
XG&$4.69e^{-01}$&--------&\green{\underline{$3.99e^{-07}$}}&\green{\underline{$4.83e^{-03}$}}&\green{\underline{$2.31e^{-10}$}}&$4.06e^{-01}$\\
Rocket&\green{\underline{$1.13e^{-12}$}}&\green{\underline{$1.12e^{-12}$}}&--------&\red{$\mathit{6.07e^{-04}}$}&\green{\underline{$3.41e^{-05}$}}&\red{$\mathit{8.08e^{-08}}$}\\
ANN&$1.34e^{-02}$&\green{\underline{$5.64e^{-05}$}}&\red{$\mathit{1.07e^{-12}}$}&--------&\green{\underline{$3.52e^{-09}$}}&\red{$\mathit{9.34e^{-03}}$}\\
1D CNN&\green{\underline{$5.65e^{-11}$}}&\green{\underline{$1.04e^{-10}$}}&$1.54e^{-01}$&\green{\underline{$1.13e^{-09}$}}&--------&\red{$\mathit{2.73e^{-11}}$}\\
CNN&\green{\underline{$1.13e^{-12}$}}&\green{\underline{$1.12e^{-12}$}}&$1.00e^{+00}$&\green{\underline{$1.07e^{-12}$}}&$1.54e^{-01}$&--------\\
&\multicolumn{6}{c}{Training}\\
\end{tabular}
}\label{pvalues_BMI_VEN}
\end{table}


\begin{table}[!h]
\centering
\caption{Kruskal-Wallis \textit{p}-values comparing all methods of the BMI\_HR+VEN prediction task. Above the diagonal, we have the results for the test set, and below for the training set. Significant results~($p < 0.01$) are represented in \green{\underline{green}} or \red{\textit{red}} when the method on the left is significantly better or worse, respectively.}
\resizebox{\linewidth}{!}{\begin{tabular}{c|ccccccc}
BMI\_HR+VEN&RF&XG&Rocket&ANN&1D CNN&CNN&\\
\hline
RF&--------&$8.11e^{-01}$&\green{\underline{$1.77e^{-08}$}}&\green{\underline{$2.21e^{-05}$}}&\green{\underline{$2.57e^{-11}$}}&$4.28e^{-01}$\parbox[c]{0mm}{\multirow{7}{*}{\rotatebox[origin=t]{90}{\makecell{\\Test}}}}\\
XG&$8.94e^{-01}$&--------&\green{\underline{$3.47e^{-09}$}}&\green{\underline{$2.72e^{-06}$}}&\green{\underline{$2.49e^{-11}$}}&$3.31e^{-01}$\\
Rocket&\green{\underline{$1.12e^{-12}$}}&\green{\underline{$4.28e^{-12}$}}&--------&\red{$\mathit{8.03e^{-04}}$}&\green{\underline{$2.60e^{-11}$}}&\red{$\mathit{4.55e^{-10}}$}\\
ANN&\green{\underline{$1.25e^{-03}$}}&\green{\underline{$7.09e^{-07}$}}&\red{$\mathit{1.08e^{-12}}$}&--------&\green{\underline{$2.44e^{-11}$}}&\red{$\mathit{7.22e^{-07}}$}\\
1D CNN&\red{$\mathit{2.66e^{-11}}$}&\red{$\mathit{2.67e^{-11}}$}&\red{$\mathit{1.06e^{-12}}$}&\red{$\mathit{2.57e^{-11}}$}&--------&\red{$\mathit{1.69e^{-11}}$}\\
CNN&\green{\underline{$1.12e^{-12}$}}&\green{\underline{$4.28e^{-12}$}}&$1.00e^{+00}$&\green{\underline{$1.08e^{-12}$}}&\green{\underline{$1.06e^{-12}$}}&--------\\
&\multicolumn{6}{c}{Training}\\
\end{tabular}
}\label{pvalues_BMI_HR+VEN}
\end{table}


\begin{table}[!h]
\centering
\caption{Kruskal-Wallis \textit{p}-values comparing all methods of the Age\_Siamese prediction task. Above the diagonal, we have the results for the test set, and below for the training set. Significant results~($p < 0.01$) are represented in \green{\underline{green}} or \red{\textit{red}} when the method on the left is significantly better or worse, respectively.}
\resizebox{0.75\linewidth}{!}{
\begin{tabular}{c|cccc}
Age\_Siamese&ANN&1D CNN&CNN&\\
\hline
ANN&--------&$5.11e^{-01}$&$3.13e^{-01}$\parbox[c]{0mm}{\multirow{4}{*}{\rotatebox[origin=t]{90}{\makecell{\\Test}}}}\\
1D CNN&\green{\underline{$2.10e^{-11}$}}&--------&$6.33e^{-02}$\\
CNN&\green{\underline{$1.12e^{-12}$}}&\green{\underline{$5.09e^{-06}$}}&--------\\
&\multicolumn{3}{c}{Training}\\
\end{tabular}
}
\label{pvalues_Age_Siamese}
\end{table}


\begin{table}[!h]
\centering
\caption{Kruskal-Wallis \textit{p}-values comparing all methods of the Age\_HR prediction task. Above the diagonal, we have the results for the test set, and below for the training set. Significant results~($p < 0.01$) are represented in \green{\underline{green}} or \red{\textit{red}} when the method on the left is significantly better or worse, respectively.}
\resizebox{\linewidth}{!}{\begin{tabular}{c|ccccccc}
Age\_HR&RF&XG&Rocket&ANN&1D CNN&CNN&\\
\hline
RF&--------&$6.51e^{-01}$&$8.99e^{-01}$&$3.82e^{-01}$&$1.51e^{-02}$&\red{$\mathit{6.23e^{-08}}$}\parbox[c]{0mm}{\multirow{7}{*}{\rotatebox[origin=t]{90}{\makecell{\\Test}}}}\\
XG&$1.51e^{-01}$&--------&$6.71e^{-01}$&$1.55e^{-01}$&$3.41e^{-02}$&\red{$\mathit{1.76e^{-08}}$}\\
Rocket&\green{\underline{$1.55e^{-11}$}}&\green{\underline{$1.53e^{-11}$}}&--------&$2.54e^{-01}$&\green{\underline{$2.75e^{-03}$}}&\red{$\mathit{1.55e^{-09}}$}\\
ANN&\red{$\mathit{1.18e^{-05}}$}&\red{$\mathit{2.37e^{-07}}$}&\red{$\mathit{1.08e^{-12}}$}&--------&\green{\underline{$1.35e^{-03}$}}&\red{$\mathit{4.58e^{-07}}$}\\
1D CNN&\green{\underline{$4.05e^{-09}$}}&\green{\underline{$3.04e^{-09}$}}&\red{$\mathit{5.32e^{-03}}$}&\green{\underline{$8.64e^{-12}$}}&--------&\red{$\mathit{8.02e^{-11}}$}\\
CNN&\green{\underline{$1.55e^{-11}$}}&\green{\underline{$1.53e^{-11}$}}&$1.00e^{+00}$&\green{\underline{$1.08e^{-12}$}}&\green{\underline{$5.32e^{-03}$}}&--------\\
&\multicolumn{6}{c}{Training}\\
\end{tabular}
}\label{pvalues_Age_HR}
\end{table}


\begin{table}[!h]
\centering
\caption{Kruskal-Wallis \textit{p}-values comparing all methods of the Age\_VEN prediction task. Above the diagonal, we have the results for the test set, and below for the training set. Significant results~($p < 0.01$) are represented in \green{\underline{green}} or \red{\textit{red}} when the method on the left is significantly better or worse, respectively.}
\resizebox{\linewidth}{!}{\begin{tabular}{c|ccccccc}
Age\_VEN&RF&XG&Rocket&ANN&1D CNN&CNN&\\
\hline
RF&--------&$3.25e^{-01}$&$6.64e^{-02}$&$6.57e^{-01}$&$4.39e^{-02}$&\red{$\mathit{7.97e^{-04}}$}\parbox[c]{0mm}{\multirow{7}{*}{\rotatebox[origin=t]{90}{\makecell{\\Test}}}}\\
XG&$2.71e^{-01}$&--------&\green{\underline{$1.19e^{-03}$}}&$1.16e^{-01}$&\green{\underline{$2.08e^{-04}$}}&\red{$\mathit{2.03e^{-04}}$}\\
Rocket&\green{\underline{$1.52e^{-11}$}}&\green{\underline{$3.29e^{-07}$}}&--------&$1.20e^{-01}$&$7.00e^{-01}$&\red{$\mathit{5.56e^{-08}}$}\\
ANN&$2.13e^{-01}$&$3.90e^{-02}$&\red{$\mathit{1.06e^{-12}}$}&--------&$8.67e^{-02}$&\red{$\mathit{1.68e^{-05}}$}\\
1D CNN&\green{\underline{$1.63e^{-07}$}}&\green{\underline{$4.54e^{-04}$}}&\red{$\mathit{5.36e^{-03}}$}&\green{\underline{$5.27e^{-10}$}}&--------&\red{$\mathit{6.99e^{-09}}$}\\
CNN&\green{\underline{$1.52e^{-11}$}}&\green{\underline{$3.29e^{-07}$}}&$1.00e^{+00}$&\green{\underline{$1.06e^{-12}$}}&\green{\underline{$5.36e^{-03}$}}&--------\\
&\multicolumn{6}{c}{Training}\\
\end{tabular}
}\label{pvalues_Age_VEN}
\end{table}


\begin{table}[!h]
\centering
\caption{Kruskal-Wallis \textit{p}-values comparing all methods of the Age\_HR+VEN prediction task. Above the diagonal, we have the results for the test set, and below for the training set. Significant results~($p < 0.01$) are represented in \green{\underline{green}} or \red{\textit{red}} when the method on the left is significantly better or worse, respectively.}
\resizebox{\linewidth}{!}{\begin{tabular}{c|ccccccc}
Age\_HR+VEN&RF&XG&Rocket&ANN&1D CNN&CNN&\\
\hline
RF&--------&\red{$\mathit{7.88e^{-03}}$}&\red{$\mathit{6.68e^{-03}}$}&\red{$\mathit{7.65e^{-04}}$}&$6.61e^{-02}$&\red{$\mathit{3.58e^{-04}}$}\parbox[c]{0mm}{\multirow{7}{*}{\rotatebox[origin=t]{90}{\makecell{\\Test}}}}\\
XG&\green{\underline{$1.99e^{-03}$}}&--------&$5.02e^{-01}$&$1.68e^{-01}$&$5.63e^{-01}$&$1.23e^{-01}$\\
Rocket&\green{\underline{$1.79e^{-10}$}}&\green{\underline{$3.00e^{-04}$}}&--------&$5.73e^{-01}$&$9.10e^{-02}$&$3.58e^{-01}$\\
ANN&$8.33e^{-02}$&\red{$\mathit{2.91e^{-05}}$}&\red{$\mathit{1.10e^{-12}}$}&--------&\green{\underline{$8.70e^{-03}$}}&$9.06e^{-01}$\\
1D CNN&\green{\underline{$2.25e^{-08}$}}&$1.16e^{-02}$&$7.82e^{-02}$&\green{\underline{$7.44e^{-11}$}}&--------&\red{$\mathit{5.57e^{-03}}$}\\
CNN&\green{\underline{$1.79e^{-10}$}}&\green{\underline{$3.00e^{-04}$}}&$1.00e^{+00}$&\green{\underline{$1.10e^{-12}$}}&$7.82e^{-02}$&--------\\
&\multicolumn{6}{c}{Training}\\
\end{tabular}
}\label{pvalues_Age_HR+VEN}
\end{table}

\onecolumn
\section{UCR $p$-values}
\label{sections:UCR-pvals}


\begin{table*}[!h]
\centering
\caption{Kruskal-Wallis \textit{p}-values comparing the test results of all methods on the BME prediction task.\\ Significant results~($p < 0.01$) are represented in \green{\underline{green}} or \red{\textit{red}} when the method on the left is significantly better or worse, respectively.}
\resizebox{\linewidth}{!}{
\begin{tabular}{c|ccccccccccccccc}
BME&TS-CHIEF&\makecell{HIVE-COTE\\v1.0}&ROCKET&\makecell{Inception\\Time}&STC&ResNet&\makecell{Proximity\\Forest}&WEASEL&S-BOSS&cBOSS&BOSS&RISE&TSF&Catch22&\\
\hline
CNN&$1.26e^{-01}$&\green{\underline{$1.35e^{-04}$}}&$1.20e^{-01}$&$3.81e^{-01}$&\green{\underline{$3.02e^{-11}$}}&\red{$\mathit{1.02e^{-03}}$}&\red{$\mathit{5.50e^{-03}}$}&\green{\underline{$9.16e^{-08}$}}&\green{\underline{$3.06e^{-11}$}}&\green{\underline{$3.05e^{-11}$}}&\green{\underline{$3.08e^{-11}$}}&\green{\underline{$3.07e^{-11}$}}&\green{\underline{$4.74e^{-09}$}}&\green{\underline{$3.07e^{-11}$}}\parbox[c]{0mm}{\multirow{16}{*}{\rotatebox[origin=t]{90}{\makecell{\\Test}}}}\\
TS-CHIEF&--------&\green{\underline{$2.75e^{-06}$}}&$8.53e^{-01}$&$3.24e^{-01}$&\green{\underline{$3.18e^{-11}$}}&$1.49e^{-02}$&$1.71e^{-01}$&\green{\underline{$1.10e^{-08}$}}&\green{\underline{$2.10e^{-11}$}}&\green{\underline{$2.09e^{-11}$}}&\green{\underline{$2.11e^{-11}$}}&\green{\underline{$1.99e^{-11}$}}&\green{\underline{$4.14e^{-10}$}}&\green{\underline{$2.35e^{-11}$}}\\
HIVE-COTE v1.0&--------&--------&\red{$\mathit{1.34e^{-06}}$}&\red{$\mathit{9.38e^{-06}}$}&\green{\underline{$8.76e^{-10}$}}&\red{$\mathit{5.93e^{-09}}$}&\red{$\mathit{1.28e^{-08}}$}&\green{\underline{$6.09e^{-04}$}}&\green{\underline{$8.83e^{-11}$}}&\green{\underline{$6.46e^{-11}$}}&\green{\underline{$7.61e^{-11}$}}&\green{\underline{$5.55e^{-11}$}}&\green{\underline{$1.12e^{-03}$}}&\green{\underline{$1.72e^{-10}$}}\\
ROCKET&--------&--------&--------&$3.96e^{-01}$&\green{\underline{$2.33e^{-11}$}}&\red{$\mathit{7.95e^{-03}}$}&$1.10e^{-01}$&\green{\underline{$6.19e^{-09}$}}&\green{\underline{$2.24e^{-11}$}}&\green{\underline{$2.23e^{-11}$}}&\green{\underline{$2.25e^{-11}$}}&\green{\underline{$2.24e^{-11}$}}&\green{\underline{$7.95e^{-11}$}}&\green{\underline{$2.24e^{-11}$}}\\
InceptionTime&--------&--------&--------&--------&\green{\underline{$3.19e^{-11}$}}&\red{$\mathit{9.81e^{-04}}$}&$1.66e^{-02}$&\green{\underline{$2.06e^{-08}$}}&\green{\underline{$3.06e^{-11}$}}&\green{\underline{$3.05e^{-11}$}}&\green{\underline{$3.08e^{-11}$}}&\green{\underline{$3.07e^{-11}$}}&\green{\underline{$2.45e^{-10}$}}&\green{\underline{$3.07e^{-11}$}}\\
STC&--------&--------&--------&--------&--------&\red{$\mathit{3.95e^{-12}}$}&\red{$\mathit{8.82e^{-12}}$}&$1.03e^{-02}$&\green{\underline{$4.95e^{-06}$}}&\green{\underline{$5.21e^{-09}$}}&\green{\underline{$2.19e^{-06}$}}&\green{\underline{$7.59e^{-11}$}}&\red{$\mathit{1.84e^{-04}}$}&\green{\underline{$7.67e^{-03}$}}\\
ResNet&--------&--------&--------&--------&--------&--------&$1.85e^{-01}$&\green{\underline{$1.54e^{-10}$}}&\green{\underline{$3.79e^{-12}$}}&\green{\underline{$3.78e^{-12}$}}&\green{\underline{$3.82e^{-12}$}}&\green{\underline{$3.80e^{-12}$}}&\green{\underline{$1.48e^{-11}$}}&\green{\underline{$3.81e^{-12}$}}\\
ProximityForest&--------&--------&--------&--------&--------&--------&--------&\green{\underline{$3.76e^{-10}$}}&\green{\underline{$8.97e^{-12}$}}&\green{\underline{$8.93e^{-12}$}}&\green{\underline{$9.03e^{-12}$}}&\green{\underline{$8.97e^{-12}$}}&\green{\underline{$8.78e^{-12}$}}&\green{\underline{$8.99e^{-12}$}}\\
WEASEL&--------&--------&--------&--------&--------&--------&--------&--------&\green{\underline{$4.93e^{-07}$}}&\green{\underline{$5.84e^{-09}$}}&\green{\underline{$2.38e^{-07}$}}&\green{\underline{$9.13e^{-10}$}}&$4.13e^{-01}$&\green{\underline{$4.82e^{-05}$}}\\
S-BOSS&--------&--------&--------&--------&--------&--------&--------&--------&--------&\green{\underline{$1.58e^{-04}$}}&$9.56e^{-01}$&\green{\underline{$3.90e^{-06}$}}&\red{$\mathit{1.59e^{-09}}$}&\red{$\mathit{4.65e^{-03}}$}\\
cBOSS&--------&--------&--------&--------&--------&--------&--------&--------&--------&--------&\red{$\mathit{1.76e^{-05}}$}&$8.39e^{-01}$&\red{$\mathit{1.99e^{-10}}$}&\red{$\mathit{1.05e^{-07}}$}\\
BOSS&--------&--------&--------&--------&--------&--------&--------&--------&--------&--------&--------&\green{\underline{$1.01e^{-07}$}}&\red{$\mathit{7.00e^{-10}}$}&\red{$\mathit{2.78e^{-03}}$}\\
RISE&--------&--------&--------&--------&--------&--------&--------&--------&--------&--------&--------&--------&\red{$\mathit{5.83e^{-11}}$}&\red{$\mathit{8.62e^{-10}}$}\\
TSF&--------&--------&--------&--------&--------&--------&--------&--------&--------&--------&--------&--------&--------&\green{\underline{$1.00e^{-07}$}}\\
\end{tabular}}
\label{pvalues_bme}
\end{table*}


\begin{table*}[!h]
\centering
\caption{Kruskal-Wallis \textit{p}-values comparing the test results of all methods on the Coffee prediction task.\\ Significant results~($p < 0.01$) are represented in \green{\underline{green}} or \red{\textit{red}} when the method on the left is significantly better or worse, respectively.}
\resizebox{\linewidth}{!}{
\begin{tabular}{c|ccccccccccccccc}
Coffee&TS-CHIEF&\makecell{HIVE-COTE\\v1.0}&ROCKET&\makecell{Inception\\Time}&STC&ResNet&\makecell{Proximity\\Forest}&WEASEL&S-BOSS&cBOSS&BOSS&RISE&TSF&Catch22&\\
\hline
CNN&\green{\underline{$2.53e^{-03}$}}&$2.05e^{-02}$&$1.00e^{+00}$&$3.17e^{-01}$&\green{\underline{$5.20e^{-03}$}}&$7.79e^{-02}$&$1.04e^{-02}$&\green{\underline{$2.56e^{-03}$}}&\green{\underline{$2.25e^{-05}$}}&\green{\underline{$2.53e^{-03}$}}&\green{\underline{$5.90e^{-04}$}}&\green{\underline{$5.90e^{-04}$}}&\green{\underline{$1.25e^{-03}$}}&\green{\underline{$2.32e^{-05}$}}\parbox[c]{0mm}{\multirow{16}{*}{\rotatebox[origin=t]{90}{\makecell{\\Test}}}}\\
TS-CHIEF&--------&$3.96e^{-01}$&\red{$\mathit{2.53e^{-03}}$}&$1.18e^{-02}$&$8.30e^{-01}$&$9.69e^{-02}$&$6.01e^{-01}$&$9.36e^{-01}$&$8.17e^{-02}$&$1.00e^{+00}$&$4.76e^{-01}$&$4.76e^{-01}$&$3.27e^{-01}$&$7.07e^{-02}$\\
HIVE-COTE v1.0&--------&--------&$2.05e^{-02}$&$8.49e^{-02}$&$5.29e^{-01}$&$4.28e^{-01}$&$7.49e^{-01}$&$3.68e^{-01}$&$1.50e^{-02}$&$3.96e^{-01}$&$1.40e^{-01}$&$1.37e^{-01}$&$1.38e^{-01}$&$1.34e^{-02}$\\
ROCKET&--------&--------&--------&$3.17e^{-01}$&\green{\underline{$5.20e^{-03}$}}&$7.79e^{-02}$&$1.04e^{-02}$&\green{\underline{$2.56e^{-03}$}}&\green{\underline{$2.25e^{-05}$}}&\green{\underline{$2.53e^{-03}$}}&\green{\underline{$5.90e^{-04}$}}&\green{\underline{$5.90e^{-04}$}}&\green{\underline{$1.25e^{-03}$}}&\green{\underline{$2.32e^{-05}$}}\\
InceptionTime&--------&--------&--------&--------&$2.29e^{-02}$&$3.04e^{-01}$&$4.45e^{-02}$&$1.16e^{-02}$&\green{\underline{$1.11e^{-04}$}}&$1.18e^{-02}$&\green{\underline{$2.69e^{-03}$}}&\green{\underline{$2.69e^{-03}$}}&\green{\underline{$4.33e^{-03}$}}&\green{\underline{$1.10e^{-04}$}}\\
STC&--------&--------&--------&--------&--------&$1.58e^{-01}$&$7.56e^{-01}$&$7.83e^{-01}$&$6.37e^{-02}$&$8.30e^{-01}$&$3.84e^{-01}$&$3.79e^{-01}$&$3.08e^{-01}$&$5.70e^{-02}$\\
ResNet&--------&--------&--------&--------&--------&--------&$2.65e^{-01}$&$9.07e^{-02}$&\green{\underline{$1.45e^{-03}$}}&$9.69e^{-02}$&$2.51e^{-02}$&$2.51e^{-02}$&$2.70e^{-02}$&\green{\underline{$1.34e^{-03}$}}\\
ProximityForest&--------&--------&--------&--------&--------&--------&--------&$5.59e^{-01}$&$3.18e^{-02}$&$6.01e^{-01}$&$2.40e^{-01}$&$2.36e^{-01}$&$2.13e^{-01}$&$2.80e^{-02}$\\
WEASEL&--------&--------&--------&--------&--------&--------&--------&--------&$1.08e^{-01}$&$9.36e^{-01}$&$5.39e^{-01}$&$5.32e^{-01}$&$4.21e^{-01}$&$9.42e^{-02}$\\
S-BOSS&--------&--------&--------&--------&--------&--------&--------&--------&--------&$8.17e^{-02}$&$3.35e^{-01}$&$3.44e^{-01}$&$6.26e^{-01}$&$9.03e^{-01}$\\
cBOSS&--------&--------&--------&--------&--------&--------&--------&--------&--------&--------&$4.76e^{-01}$&$4.76e^{-01}$&$3.27e^{-01}$&$7.07e^{-02}$\\
BOSS&--------&--------&--------&--------&--------&--------&--------&--------&--------&--------&--------&$9.85e^{-01}$&$8.15e^{-01}$&$2.94e^{-01}$\\
RISE&--------&--------&--------&--------&--------&--------&--------&--------&--------&--------&--------&--------&$8.15e^{-01}$&$3.06e^{-01}$\\
TSF&--------&--------&--------&--------&--------&--------&--------&--------&--------&--------&--------&--------&--------&$5.18e^{-01}$\\
\end{tabular}
}
\label{pvalues_coffee}
\end{table*}


\begin{table*}[!h]
\centering
\caption{Kruskal-Wallis \textit{p}-values comparing the test results of all methods on the Earthquakes prediction task.\\ Significant results~($p < 0.01$) are represented in \green{\underline{green}} or \red{\textit{red}} when the method on the left is significantly better or worse, respectively.}
\resizebox{\linewidth}{!}{
\begin{tabular}{c|ccccccccccccccc}
Earthquakes&TS-CHIEF&\makecell{HIVE-COTE\\v1.0}&ROCKET&\makecell{Inception\\Time}&STC&ResNet&\makecell{Proximity\\Forest}&WEASEL&S-BOSS&cBOSS&BOSS&RISE&TSF&Catch22&\\
\hline
CNN&\green{\underline{$1.47e^{-03}$}}&\green{\underline{$1.54e^{-03}$}}&\green{\underline{$1.59e^{-03}$}}&\green{\underline{$1.38e^{-05}$}}&\green{\underline{$4.22e^{-04}$}}&\green{\underline{$2.46e^{-08}$}}&\green{\underline{$7.31e^{-03}$}}&\green{\underline{$1.54e^{-03}$}}&\green{\underline{$1.54e^{-03}$}}&\green{\underline{$1.54e^{-03}$}}&\green{\underline{$1.49e^{-03}$}}&\green{\underline{$1.47e^{-03}$}}&\green{\underline{$1.72e^{-03}$}}&\green{\underline{$3.81e^{-04}$}}\parbox[c]{0mm}{\multirow{16}{*}{\rotatebox[origin=t]{90}{\makecell{\\Test}}}}\\
TS-CHIEF&--------&$7.79e^{-02}$&$3.17e^{-01}$&\green{\underline{$5.40e^{-06}$}}&\green{\underline{$3.96e^{-06}$}}&\green{\underline{$2.53e^{-10}$}}&$3.27e^{-02}$&$7.79e^{-02}$&$7.79e^{-02}$&$7.79e^{-02}$&\green{\underline{$2.56e^{-03}$}}&$1.00e^{+00}$&\red{$\mathit{1.50e^{-07}}$}&\green{\underline{$5.73e^{-07}$}}\\
HIVE-COTE v1.0&--------&--------&$4.54e^{-02}$&\green{\underline{$3.81e^{-05}$}}&\green{\underline{$2.42e^{-04}$}}&\green{\underline{$1.09e^{-09}$}}&$1.92e^{-02}$&$1.00e^{+00}$&$1.00e^{+00}$&$1.00e^{+00}$&$9.07e^{-02}$&$7.79e^{-02}$&\red{$\mathit{2.69e^{-07}}$}&\green{\underline{$1.06e^{-05}$}}\\
ROCKET&--------&--------&--------&\green{\underline{$5.54e^{-06}$}}&\green{\underline{$3.43e^{-06}$}}&\green{\underline{$3.28e^{-10}$}}&$5.32e^{-02}$&$4.54e^{-02}$&$4.54e^{-02}$&$4.54e^{-02}$&\green{\underline{$1.76e^{-03}$}}&$3.17e^{-01}$&\red{$\mathit{1.71e^{-06}}$}&\green{\underline{$5.64e^{-07}$}}\\
InceptionTime&--------&--------&--------&--------&$3.43e^{-02}$&\green{\underline{$7.25e^{-03}$}}&\red{$\mathit{1.84e^{-04}}$}&\red{$\mathit{3.81e^{-05}}$}&\red{$\mathit{3.81e^{-05}}$}&\red{$\mathit{3.81e^{-05}}$}&\red{$\mathit{6.29e^{-04}}$}&\red{$\mathit{5.40e^{-06}}$}&\red{$\mathit{4.38e^{-06}}$}&$2.92e^{-01}$\\
STC&--------&--------&--------&--------&--------&\green{\underline{$7.84e^{-07}$}}&\red{$\mathit{4.54e^{-04}}$}&\red{$\mathit{2.42e^{-04}}$}&\red{$\mathit{2.42e^{-04}}$}&\red{$\mathit{2.42e^{-04}}$}&$2.55e^{-02}$&\red{$\mathit{3.96e^{-06}}$}&\red{$\mathit{1.96e^{-07}}$}&$1.13e^{-01}$\\
ResNet&--------&--------&--------&--------&--------&--------&\red{$\mathit{5.97e^{-08}}$}&\red{$\mathit{1.09e^{-09}}$}&\red{$\mathit{1.09e^{-09}}$}&\red{$\mathit{1.09e^{-09}}$}&\red{$\mathit{8.98e^{-09}}$}&\red{$\mathit{2.53e^{-10}}$}&\red{$\mathit{1.55e^{-09}}$}&\red{$\mathit{1.82e^{-05}}$}\\
ProximityForest&--------&--------&--------&--------&--------&--------&--------&$1.92e^{-02}$&$1.92e^{-02}$&$1.92e^{-02}$&\green{\underline{$6.54e^{-03}$}}&$3.27e^{-02}$&$6.34e^{-01}$&\green{\underline{$1.73e^{-04}$}}\\
WEASEL&--------&--------&--------&--------&--------&--------&--------&--------&$1.00e^{+00}$&$1.00e^{+00}$&$9.07e^{-02}$&$7.79e^{-02}$&\red{$\mathit{2.69e^{-07}}$}&\green{\underline{$1.06e^{-05}$}}\\
S-BOSS&--------&--------&--------&--------&--------&--------&--------&--------&--------&$1.00e^{+00}$&$9.07e^{-02}$&$7.79e^{-02}$&\red{$\mathit{2.69e^{-07}}$}&\green{\underline{$1.06e^{-05}$}}\\
cBOSS&--------&--------&--------&--------&--------&--------&--------&--------&--------&--------&$9.07e^{-02}$&$7.79e^{-02}$&\red{$\mathit{2.69e^{-07}}$}&\green{\underline{$1.06e^{-05}$}}\\
BOSS&--------&--------&--------&--------&--------&--------&--------&--------&--------&--------&--------&\red{$\mathit{2.56e^{-03}}$}&\red{$\mathit{3.59e^{-07}}$}&\green{\underline{$6.30e^{-04}$}}\\
RISE&--------&--------&--------&--------&--------&--------&--------&--------&--------&--------&--------&--------&\red{$\mathit{1.50e^{-07}}$}&\green{\underline{$5.73e^{-07}$}}\\
TSF&--------&--------&--------&--------&--------&--------&--------&--------&--------&--------&--------&--------&--------&\green{\underline{$1.26e^{-07}$}}\\
\end{tabular}
}
\label{pvalues_earthquakes}
\end{table*}


\begin{table*}[!h]
\centering
\caption{Kruskal-Wallis \textit{p}-values comparing the test results of all methods on the Plane prediction task.\\ Significant results~($p < 0.01$) are represented in \green{\underline{green}} or \red{\textit{red}} when the method on the left is significantly better or worse, respectively.}
\resizebox{\linewidth}{!}{
\begin{tabular}{c|ccccccccccccccc}
Plane&TS-CHIEF&\makecell{HIVE-COTE\\v1.0}&ROCKET&\makecell{Inception\\Time}&STC&ResNet&\makecell{Proximity\\Forest}&WEASEL&S-BOSS&cBOSS&BOSS&RISE&TSF&Catch22&\\
\hline
CNN&$1.00e^{+00}$&$1.00e^{+00}$&$1.00e^{+00}$&$1.04e^{-02}$&$7.79e^{-02}$&$1.00e^{+00}$&$1.00e^{+00}$&\green{\underline{$2.53e^{-03}$}}&\green{\underline{$5.16e^{-03}$}}&$1.00e^{+00}$&$1.03e^{-02}$&\green{\underline{$2.59e^{-03}$}}&\green{\underline{$5.97e^{-04}$}}&\green{\underline{$3.01e^{-09}$}}\parbox[c]{0mm}{\multirow{16}{*}{\rotatebox[origin=t]{90}{\makecell{\\Test}}}}\\
TS-CHIEF&--------&$1.00e^{+00}$&$1.00e^{+00}$&$1.04e^{-02}$&$7.79e^{-02}$&$1.00e^{+00}$&$1.00e^{+00}$&\green{\underline{$2.53e^{-03}$}}&\green{\underline{$5.16e^{-03}$}}&$1.00e^{+00}$&$1.03e^{-02}$&\green{\underline{$2.59e^{-03}$}}&\green{\underline{$5.97e^{-04}$}}&\green{\underline{$3.01e^{-09}$}}\\
HIVE-COTE v1.0&--------&--------&$1.00e^{+00}$&$1.04e^{-02}$&$7.79e^{-02}$&$1.00e^{+00}$&$1.00e^{+00}$&\green{\underline{$2.53e^{-03}$}}&\green{\underline{$5.16e^{-03}$}}&$1.00e^{+00}$&$1.03e^{-02}$&\green{\underline{$2.59e^{-03}$}}&\green{\underline{$5.97e^{-04}$}}&\green{\underline{$3.01e^{-09}$}}\\
ROCKET&--------&--------&--------&$1.04e^{-02}$&$7.79e^{-02}$&$1.00e^{+00}$&$1.00e^{+00}$&\green{\underline{$2.53e^{-03}$}}&\green{\underline{$5.16e^{-03}$}}&$1.00e^{+00}$&$1.03e^{-02}$&\green{\underline{$2.59e^{-03}$}}&\green{\underline{$5.97e^{-04}$}}&\green{\underline{$3.01e^{-09}$}}\\
InceptionTime&--------&--------&--------&--------&$2.21e^{-01}$&$1.04e^{-02}$&$1.04e^{-02}$&$4.40e^{-01}$&$9.91e^{-01}$&$1.04e^{-02}$&$7.92e^{-01}$&$6.48e^{-01}$&$3.42e^{-01}$&\green{\underline{$9.20e^{-05}$}}\\
STC&--------&--------&--------&--------&--------&$7.79e^{-02}$&$7.79e^{-02}$&$5.40e^{-02}$&$1.68e^{-01}$&$7.79e^{-02}$&$2.81e^{-01}$&$7.92e^{-02}$&$1.29e^{-02}$&\green{\underline{$1.44e^{-07}$}}\\
ResNet&--------&--------&--------&--------&--------&--------&$1.00e^{+00}$&\green{\underline{$2.53e^{-03}$}}&\green{\underline{$5.16e^{-03}$}}&$1.00e^{+00}$&$1.03e^{-02}$&\green{\underline{$2.59e^{-03}$}}&\green{\underline{$5.97e^{-04}$}}&\green{\underline{$3.01e^{-09}$}}\\
ProximityForest&--------&--------&--------&--------&--------&--------&--------&\green{\underline{$2.53e^{-03}$}}&\green{\underline{$5.16e^{-03}$}}&$1.00e^{+00}$&$1.03e^{-02}$&\green{\underline{$2.59e^{-03}$}}&\green{\underline{$5.97e^{-04}$}}&\green{\underline{$3.01e^{-09}$}}\\
WEASEL&--------&--------&--------&--------&--------&--------&--------&--------&$3.89e^{-01}$&\red{$\mathit{2.53e^{-03}}$}&$2.71e^{-01}$&$6.91e^{-01}$&$8.34e^{-01}$&\green{\underline{$4.66e^{-03}$}}\\
S-BOSS&--------&--------&--------&--------&--------&--------&--------&--------&--------&\red{$\mathit{5.16e^{-03}}$}&$7.55e^{-01}$&$6.11e^{-01}$&$1.28e^{-01}$&\green{\underline{$7.64e^{-06}$}}\\
cBOSS&--------&--------&--------&--------&--------&--------&--------&--------&--------&--------&$1.03e^{-02}$&\green{\underline{$2.59e^{-03}$}}&\green{\underline{$5.97e^{-04}$}}&\green{\underline{$3.01e^{-09}$}}\\
BOSS&--------&--------&--------&--------&--------&--------&--------&--------&--------&--------&--------&$4.28e^{-01}$&$8.16e^{-02}$&\green{\underline{$3.14e^{-06}$}}\\
RISE&--------&--------&--------&--------&--------&--------&--------&--------&--------&--------&--------&--------&$4.59e^{-01}$&\green{\underline{$1.95e^{-04}$}}\\
TSF&--------&--------&--------&--------&--------&--------&--------&--------&--------&--------&--------&--------&--------&\green{\underline{$7.44e^{-03}$}}\\
\end{tabular}
}
\label{pvalues_plane}
\end{table*}


\begin{table*}[!h]
\centering
\caption{Kruskal-Wallis \textit{p}-values comparing the test results of all methods on the SmoothSubspace prediction task. \\Significant results~($p < 0.01$) are represented in \green{\underline{green}} or \red{\textit{red}} when the method on the left is significantly better or worse, respectively.}
\resizebox{\linewidth}{!}{
\begin{tabular}{c|ccccccccccccccc}
SmoothSubspace&TS-CHIEF&\makecell{HIVE-COTE\\v1.0}&ROCKET&\makecell{Inception\\Time}&STC&ResNet&\makecell{Proximity\\Forest}&WEASEL&S-BOSS&cBOSS&BOSS&RISE&TSF&Catch22&\\
\hline
CNN&\green{\underline{$9.89e^{-03}$}}&\green{\underline{$2.53e^{-09}$}}&\green{\underline{$5.03e^{-12}$}}&\green{\underline{$6.10e^{-12}$}}&\green{\underline{$4.99e^{-12}$}}&\green{\underline{$2.49e^{-05}$}}&$3.72e^{-02}$&\green{\underline{$5.03e^{-12}$}}&\green{\underline{$5.21e^{-12}$}}&\green{\underline{$5.31e^{-12}$}}&\green{\underline{$5.13e^{-12}$}}&\green{\underline{$5.16e^{-12}$}}&\green{\underline{$8.93e^{-10}$}}&\green{\underline{$5.09e^{-12}$}}\parbox[c]{0mm}{\multirow{16}{*}{\rotatebox[origin=t]{90}{\makecell{\\Test}}}}\\
TS-CHIEF&--------&\green{\underline{$1.89e^{-06}$}}&\green{\underline{$1.04e^{-10}$}}&\green{\underline{$1.53e^{-09}$}}&\green{\underline{$2.54e^{-11}$}}&$2.31e^{-02}$&$5.83e^{-01}$&\green{\underline{$2.43e^{-11}$}}&\green{\underline{$2.50e^{-11}$}}&\green{\underline{$2.54e^{-11}$}}&\green{\underline{$2.47e^{-11}$}}&\green{\underline{$2.48e^{-11}$}}&\green{\underline{$3.84e^{-08}$}}&\green{\underline{$2.45e^{-11}$}}\\
HIVE-COTE v1.0&--------&--------&\green{\underline{$1.49e^{-04}$}}&$3.11e^{-01}$&\green{\underline{$1.14e^{-10}$}}&\red{$\mathit{7.71e^{-03}}$}&\red{$\mathit{3.46e^{-07}}$}&\green{\underline{$5.10e^{-11}$}}&\green{\underline{$5.25e^{-11}$}}&\green{\underline{$5.33e^{-11}$}}&\green{\underline{$5.18e^{-11}$}}&\green{\underline{$5.20e^{-11}$}}&$7.18e^{-01}$&\green{\underline{$5.15e^{-11}$}}\\
ROCKET&--------&--------&--------&\red{$\mathit{1.79e^{-04}}$}&\green{\underline{$5.99e^{-09}$}}&\red{$\mathit{9.71e^{-08}}$}&\red{$\mathit{3.73e^{-11}}$}&\green{\underline{$5.09e^{-11}$}}&\green{\underline{$5.24e^{-11}$}}&\green{\underline{$5.32e^{-11}$}}&\green{\underline{$5.17e^{-11}$}}&\green{\underline{$5.19e^{-11}$}}&\red{$\mathit{1.24e^{-05}}$}&\green{\underline{$5.14e^{-11}$}}\\
InceptionTime&--------&--------&--------&--------&\green{\underline{$7.86e^{-11}$}}&\red{$\mathit{1.02e^{-04}}$}&\red{$\mathit{2.56e^{-10}}$}&\green{\underline{$4.07e^{-11}$}}&\green{\underline{$4.20e^{-11}$}}&\green{\underline{$4.26e^{-11}$}}&\green{\underline{$4.14e^{-11}$}}&\green{\underline{$4.16e^{-11}$}}&$3.85e^{-02}$&\green{\underline{$4.11e^{-11}$}}\\
STC&--------&--------&--------&--------&--------&\red{$\mathit{6.29e^{-11}}$}&\red{$\mathit{1.88e^{-11}}$}&\green{\underline{$1.48e^{-10}$}}&\green{\underline{$5.72e^{-11}$}}&\green{\underline{$5.81e^{-11}$}}&\green{\underline{$5.65e^{-11}$}}&\green{\underline{$5.97e^{-11}$}}&\red{$\mathit{1.10e^{-10}}$}&\green{\underline{$8.95e^{-11}$}}\\
ResNet&--------&--------&--------&--------&--------&--------&\red{$\mathit{6.32e^{-03}}$}&\green{\underline{$4.41e^{-11}$}}&\green{\underline{$4.54e^{-11}$}}&\green{\underline{$4.61e^{-11}$}}&\green{\underline{$4.48e^{-11}$}}&\green{\underline{$4.50e^{-11}$}}&\green{\underline{$2.60e^{-03}$}}&\green{\underline{$4.45e^{-11}$}}\\
ProximityForest&--------&--------&--------&--------&--------&--------&--------&\green{\underline{$1.89e^{-11}$}}&\green{\underline{$1.95e^{-11}$}}&\green{\underline{$1.99e^{-11}$}}&\green{\underline{$1.93e^{-11}$}}&\green{\underline{$1.94e^{-11}$}}&\green{\underline{$1.82e^{-08}$}}&\green{\underline{$1.91e^{-11}$}}\\
WEASEL&--------&--------&--------&--------&--------&--------&--------&--------&\green{\underline{$5.76e^{-11}$}}&\green{\underline{$5.85e^{-11}$}}&\green{\underline{$5.68e^{-11}$}}&$3.24e^{-01}$&\red{$\mathit{4.88e^{-11}}$}&$7.37e^{-01}$\\
S-BOSS&--------&--------&--------&--------&--------&--------&--------&--------&--------&$6.50e^{-02}$&$5.22e^{-01}$&\red{$\mathit{5.88e^{-11}}$}&\red{$\mathit{5.02e^{-11}}$}&\red{$\mathit{5.81e^{-11}}$}\\
cBOSS&--------&--------&--------&--------&--------&--------&--------&--------&--------&--------&$2.87e^{-02}$&\red{$\mathit{5.96e^{-11}}$}&\red{$\mathit{5.10e^{-11}}$}&\red{$\mathit{5.90e^{-11}}$}\\
BOSS&--------&--------&--------&--------&--------&--------&--------&--------&--------&--------&--------&\red{$\mathit{5.80e^{-11}}$}&\red{$\mathit{4.96e^{-11}}$}&\red{$\mathit{5.73e^{-11}}$}\\
RISE&--------&--------&--------&--------&--------&--------&--------&--------&--------&--------&--------&--------&\red{$\mathit{4.98e^{-11}}$}&$3.65e^{-01}$\\
TSF&--------&--------&--------&--------&--------&--------&--------&--------&--------&--------&--------&--------&--------&\green{\underline{$4.93e^{-11}$}}\\
\end{tabular}
}
\label{pvalues_smoothsubspace}
\end{table*}


\begin{table*}[!h]
\centering
\caption{Kruskal-Wallis \textit{p}-values comparing the test results of all methods on the Trace prediction task.\\ Significant results~($p < 0.01$) are represented in \green{\underline{green}} or \red{\textit{red}} when the method on the left is significantly better or worse, respectively.}
\resizebox{\linewidth}{!}{
\begin{tabular}{c|ccccccccccccccc}
Trace&TS-CHIEF&\makecell{HIVE-COTE\\v1.0}&ROCKET&\makecell{Inception\\Time}&STC&ResNet&\makecell{Proximity\\Forest}&WEASEL&S-BOSS&cBOSS&BOSS&RISE&TSF&Catch22&\\
\hline
CNN&$1.00e^{+00}$&$1.00e^{+00}$&$1.00e^{+00}$&$1.00e^{+00}$&$1.00e^{+00}$&$1.00e^{+00}$&$1.00e^{+00}$&$1.00e^{+00}$&$1.00e^{+00}$&$1.00e^{+00}$&$1.00e^{+00}$&\green{\underline{$3.63e^{-09}$}}&\green{\underline{$2.61e^{-03}$}}&$3.17e^{-01}$\parbox[c]{0mm}{\multirow{16}{*}{\rotatebox[origin=t]{90}{\makecell{\\Test}}}}\\
TS-CHIEF&--------&$1.00e^{+00}$&$1.00e^{+00}$&$1.00e^{+00}$&$1.00e^{+00}$&$1.00e^{+00}$&$1.00e^{+00}$&$1.00e^{+00}$&$1.00e^{+00}$&$1.00e^{+00}$&$1.00e^{+00}$&\green{\underline{$3.63e^{-09}$}}&\green{\underline{$2.61e^{-03}$}}&$3.17e^{-01}$\\
HIVE-COTE v1.0&--------&--------&$1.00e^{+00}$&$1.00e^{+00}$&$1.00e^{+00}$&$1.00e^{+00}$&$1.00e^{+00}$&$1.00e^{+00}$&$1.00e^{+00}$&$1.00e^{+00}$&$1.00e^{+00}$&\green{\underline{$3.63e^{-09}$}}&\green{\underline{$2.61e^{-03}$}}&$3.17e^{-01}$\\
ROCKET&--------&--------&--------&$1.00e^{+00}$&$1.00e^{+00}$&$1.00e^{+00}$&$1.00e^{+00}$&$1.00e^{+00}$&$1.00e^{+00}$&$1.00e^{+00}$&$1.00e^{+00}$&\green{\underline{$3.63e^{-09}$}}&\green{\underline{$2.61e^{-03}$}}&$3.17e^{-01}$\\
InceptionTime&--------&--------&--------&--------&$1.00e^{+00}$&$1.00e^{+00}$&$1.00e^{+00}$&$1.00e^{+00}$&$1.00e^{+00}$&$1.00e^{+00}$&$1.00e^{+00}$&\green{\underline{$3.63e^{-09}$}}&\green{\underline{$2.61e^{-03}$}}&$3.17e^{-01}$\\
STC&--------&--------&--------&--------&--------&$1.00e^{+00}$&$1.00e^{+00}$&$1.00e^{+00}$&$1.00e^{+00}$&$1.00e^{+00}$&$1.00e^{+00}$&\green{\underline{$3.63e^{-09}$}}&\green{\underline{$2.61e^{-03}$}}&$3.17e^{-01}$\\
ResNet&--------&--------&--------&--------&--------&--------&$1.00e^{+00}$&$1.00e^{+00}$&$1.00e^{+00}$&$1.00e^{+00}$&$1.00e^{+00}$&\green{\underline{$3.63e^{-09}$}}&\green{\underline{$2.61e^{-03}$}}&$3.17e^{-01}$\\
ProximityForest&--------&--------&--------&--------&--------&--------&--------&$1.00e^{+00}$&$1.00e^{+00}$&$1.00e^{+00}$&$1.00e^{+00}$&\green{\underline{$3.63e^{-09}$}}&\green{\underline{$2.61e^{-03}$}}&$3.17e^{-01}$\\
WEASEL&--------&--------&--------&--------&--------&--------&--------&--------&$1.00e^{+00}$&$1.00e^{+00}$&$1.00e^{+00}$&\green{\underline{$3.63e^{-09}$}}&\green{\underline{$2.61e^{-03}$}}&$3.17e^{-01}$\\
S-BOSS&--------&--------&--------&--------&--------&--------&--------&--------&--------&$1.00e^{+00}$&$1.00e^{+00}$&\green{\underline{$3.63e^{-09}$}}&\green{\underline{$2.61e^{-03}$}}&$3.17e^{-01}$\\
cBOSS&--------&--------&--------&--------&--------&--------&--------&--------&--------&--------&$1.00e^{+00}$&\green{\underline{$3.63e^{-09}$}}&\green{\underline{$2.61e^{-03}$}}&$3.17e^{-01}$\\
BOSS&--------&--------&--------&--------&--------&--------&--------&--------&--------&--------&--------&\green{\underline{$3.63e^{-09}$}}&\green{\underline{$2.61e^{-03}$}}&$3.17e^{-01}$\\
RISE&--------&--------&--------&--------&--------&--------&--------&--------&--------&--------&--------&--------&\red{$\mathit{1.25e^{-03}}$}&\red{$\mathit{1.27e^{-08}}$}\\
TSF&--------&--------&--------&--------&--------&--------&--------&--------&--------&--------&--------&--------&--------&$1.02e^{-02}$\\
\end{tabular}
}
\label{pvalues_trace}
\end{table*}

\onecolumn
\section{UCR problems}
\label{sections:UCR-problems}


\footnotesize
\begin{longtable}{c|cc}
\caption{Ranking by dataset from the set of algorithms present in UCR and our CNN methodology on the selected 98 UCR problems.} \\

&$1^{st}$&$2^{nd}$\\ 
\hline
ACSF1&ResNet (0.9300)&Over 2 methods (0.9200)\\ 
Adiac&InceptionTime (0.8517)&ResNet (0.8491)\\ 
ArrowHead&TS-CHIEF \& ProximityForest (0.9486)&\\ 
Beef&STC (1.0000)&WEASEL \& RISE (0.9333)\\ 
BeetleFly&Over 2 methods (1.0000)&\\ 
BirdChicken&Over 2 methods (1.0000)&\\ 
BME&\green{\underline{Over 2 methods}} (1.0000)&\\ 
Car&InceptionTime (0.9667)&Over 2 methods (0.9500)\\ 
CBF&Over 2 methods (1.0000)&\\ 
Chinatown&InceptionTime (0.9883)&ResNet (0.9855)\\ 
ChlorineConcentration&InceptionTime (0.8930)&ResNet (0.8703)\\ 
CinCECGTorso&HIVE-COTE v1.0 (1.0000)&WEASEL (0.9993)\\ 
Coffee&\green{\underline{Over 2 methods}} (1.0000)&\\ 
Computers&InceptionTime (0.9040)&ResNet (0.8920)\\ 
Crop&InceptionTime (0.7968)&HIVE-COTE v1.0 (0.7730)\\ 
DiatomSizeReduction&InceptionTime (1.0000)&ROCKET \& BOSS (0.9935)\\ 
DistalPhalanxOutlineAgeGroup&S-BOSS (0.9065)&Over 2 methods (0.8993)\\ 
DistalPhalanxOutlineCorrect&ROCKET \& ProximityForest (0.8696)&\\ 
DistalPhalanxTW&Over 2 methods (0.7410)&\\ 
Earthquakes&\green{\underline{CNN}} (0.7770)&ProximityForest (0.7698)\\ 
ECG200&ResNet (0.9500)&ROCKET \& InceptionTime (0.9400)\\ 
ECG5000&TS-CHIEF (0.9520)&WEASEL (0.9513)\\ 
ECGFiveDays&Over 2 methods (1.0000)&\\ 
ElectricDevices&ROCKET (0.9046)&InceptionTime (0.9042)\\ 
EthanolLevel&ResNet (0.9080)&InceptionTime (0.9040)\\ 
FaceAll&InceptionTime \& ResNet (0.9941)&\\ 
FaceFour&Over 2 methods (1.0000)&\\ 
FacesUCR&InceptionTime (0.9859)&TS-CHIEF (0.9839)\\ 
FiftyWords&TS-CHIEF \& InceptionTime (0.8659)&\\ 
Fish&Over 2 methods (1.0000)&\\ 
FordA&WEASEL (0.9765)&InceptionTime (0.9689)\\ 
FreezerRegularTrain&HIVE-COTE v1.0 \& STC (1.0000)&\\ 
FreezerSmallTrain&HIVE-COTE v1.0 \& STC (1.0000)&\\ 
GunPoint&Over 2 methods (1.0000)&\\ 
GunPointAgeSpan&Over 2 methods (1.0000)&\\ 
GunPointMaleVersusFemale&Over 2 methods (1.0000)&\\ 
GunPointOldVersusYoung&Over 2 methods (1.0000)&\\ 
Ham&ProximityForest (1.0000)&ROCKET (0.9968)\\ 
Haptics&HIVE-COTE v1.0 (0.9486)&ResNet (0.9351)\\ 
Herring&Catch22 (0.9054)&S-BOSS (0.7344)\\ 
HouseTwenty&Over 2 methods (0.9916)&\\ 
InlineSkate&HIVE-COTE v1.0 (1.0000)&Over 2 methods (0.9916)\\ 
InsectEPGRegularTrain&Over 2 methods (1.0000)&\\ 
InsectEPGSmallTrain&Over 2 methods (1.0000)&\\ 
ItalyPowerDemand&TS-CHIEF \& InceptionTime (0.9728)&\\ 
LargeKitchenAppliances&InceptionTime (0.9733)&ResNet (0.9718)\\ 
Lightning2&ResNet (0.9760)&HIVE-COTE v1.0 (0.9520)\\ 
Lightning7&Catch22 (0.9253)&ProximityForest (0.9180)\\ 
Mallat&TS-CHIEF (0.9979)&WEASEL (0.9851)\\ 
Meat&Over 2 methods (1.0000)&\\ 
MedicalImages&Over 2 methods (1.0000)&\\ 
MiddlePhalanxOutlineAgeGroup&Catch22 (0.9833)&ROCKET (0.8289)\\ 
MiddlePhalanxOutlineCorrect&STC (0.8694)&InceptionTime (0.8660)\\ 
MiddlePhalanxTW&ROCKET (0.8660)&ProximityForest (0.8625)\\ 
MixedShapesRegularTrain&TS-CHIEF (0.9810)&InceptionTime \& WEASEL (0.9753)\\ 
MixedShapesSmallTrain&ResNet (0.9790)&HIVE-COTE v1.0 (0.9769)\\ 
MoteStrain&TS-CHIEF (0.9688)&HIVE-COTE v1.0 (0.9662)\\ 
OliveOil&Over 2 methods (1.0000)&\\ 
OSULeaf&ROCKET \& ProximityForest (1.0000)&\\ 
PhalangesOutlinesCorrect&HIVE-COTE v1.0 (0.9623)&ROCKET (0.9587)\\ 
Phoneme&HIVE-COTE v1.0 \& ResNet (0.9667)&\\ 
PigAirwayPressure&STC (0.9952)&HIVE-COTE v1.0 \& ResNet (0.9876)\\ 
PigArtPressure&S-BOSS (1.0000)&Over 2 methods (0.9952)\\ 
PigCVP&TS-CHIEF (0.9856)&S-BOSS (0.9808)\\ 
Plane&\green{\underline{Over 2 methods}} (1.0000)&\\ 
ProximalPhalanxOutlineAgeGroup&ProximityForest (1.0000)&ROCKET (0.9889)\\ 
ProximalPhalanxOutlineCorrect&HIVE-COTE v1.0 \& ResNet (1.0000)&\\ 
ProximalPhalanxTW&HIVE-COTE v1.0 \& Catch22 (1.0000)&\\ 
RefrigerationDevices&Catch22 (0.9389)&HIVE-COTE v1.0 (0.8878)\\ 
Rock&STC (0.9600)&WEASEL (0.9400)\\ 
ScreenType&ROCKET (0.9000)&Catch22 (0.8660)\\ 
SemgHandGenderCh2&TS-CHIEF (0.9683)&TSF (0.9650)\\ 
SemgHandMovementCh2&ProximityForest (0.9800)&ROCKET (0.9500)\\ 
SemgHandSubjectCh2&TSF (0.9533)&TS-CHIEF (0.9511)\\ 
ShapeletSim&Over 2 methods (1.0000)&\\ 
ShapesAll&ROCKET (1.0000)&InceptionTime (0.9600)\\ 
SmallKitchenAppliances&HIVE-COTE v1.0 (0.9689)&ROCKET (0.9417)\\ 
SmoothSubspace&\green{\underline{Over 2 methods}} (1.0000)&\\ 
SonyAIBORobotSurface1&ProximityForest \& Catch22 (1.0000)&\\ 
SonyAIBORobotSurface2&ROCKET (0.9817)&InceptionTime (0.9748)\\ 
StarLightCurves&HIVE-COTE v1.0 \& ResNet (1.0000)&\\ 
Strawberry&WEASEL (0.9919)&Over 2 methods (0.9892)\\ 
SwedishLeaf&ROCKET \& ProximityForest (0.9919)&\\ 
Symbols&TS-CHIEF (0.9879)&InceptionTime (0.9859)\\ 
SyntheticControl&Over 2 methods (1.0000)&\\ 
ToeSegmentation1&ROCKET \& ProximityForest (1.0000)&\\ 
ToeSegmentation2&TS-CHIEF \& InceptionTime (0.9923)&\\ 
Trace&\green{\underline{Over 2 methods}} (1.0000)&\\ 
TwoLeadECG&Over 2 methods (1.0000)&\\ 
TwoPatterns&Over 2 methods (1.0000)&\\ 
UMD&Over 2 methods (1.0000)&\\ 
UWaveGestureLibraryAll&Over 2 methods (1.0000)&\\ 
Wafer&Over 2 methods (1.0000)&\\ 
Wine&Over 2 methods (1.0000)&\\ 
WordSynonyms&ROCKET \& ProximityForest (1.0000)&\\ 
Worms&HIVE-COTE v1.0 \& ResNet (1.0000)&\\ 
WormsTwoClass&HIVE-COTE v1.0 \& ResNet (1.0000)&\\ 
Yoga&cBOSS (0.9373)&InceptionTime (0.9337)\\ 
\label{giga_table}
\end{longtable}
\normalsize

\twocolumn

\end{document}